\documentclass{article}

\PassOptionsToPackage{round}{natbib}

 \usepackage[preprint]{neurips_2026}

\usepackage[utf8]{inputenc} 
\usepackage[T1]{fontenc}    
\usepackage{hyperref}       
\usepackage{url}            
\usepackage{booktabs}       
\usepackage{amsfonts}       
\usepackage{nicefrac}       
\usepackage{microtype}      
\usepackage{xcolor}         
\usepackage{float}
\usepackage{amsmath}
\usepackage{graphicx}
\usepackage{rotating}
\graphicspath{{figures/}}   

\title{
Physics-Informed Neural Networks for Predicting Nitrous Oxide Flux%
}

\author{%
  Freddy Yu\thanks{Alternate email: \texttt{freddyyu99@gmail.com}} \\
  Harvard University\\
  \texttt{freddyyu@college.harvard.edu} \\
  \AND
  Jashanjeet Kaur Dhaliwal \\
  Department of Biosystems Engineering and Soil Science \\
  University of Tennessee \\
  \texttt{jdhaliwa@utk.edu} \\
  \AND
  Subhadeep Chakraborty \\
  Department of Mechanical and Aerospace Engineering \\
  University of Tennessee \\
  \texttt{schakrab@utk.edu} \\
}

\begin{document}

\maketitle

\begin{abstract}
Nitrous oxide (N$_2$O) is the dominant ozone-depleting substance emitted in the 21st century, and the third largest contributor to anthropogenic greenhouse gases due to its high potency and long atmospheric lifetime, with more than 70\% of N$_2$O emissions occurring as a result of agricultural processes. Current approaches to predicting N$_2$O flux emissions include process-based models such as DayCent and Cycles, as well as classical AI models, but the application of Physics-Informed Neural Networks (PINNs) to predicting N$_2$O flux emissions is largely underexplored. Our paper draws upon the mechanistic equations that underlie the DayCent family of process-based models to construct a rigorously derived, literature-traceable physics residual. We then build and train an MLP-based PINN on a multi-site agricultural dataset spanning four geographically distinct US agricultural sites. Across all tested values of the physics loss weighting hyperparameter $\lambda$, our PINN consistently and substantially outperformed uncalibrated Cycles simulation (R$^2=0.01$), with our MLP baseline achieving mean R$^2=0.411$ across ten random seeds. Physics constraints consistently degrade model performance in holdout validation, with marginal degradation at low $\lambda$ and significant degradation at high $\lambda$, but consistently improve model performance and reduce performance variability in leave-one-site-out validation. This suggests that physics constraints sacrifice in-distribution accuracy for out-of-distribution robustness, anchoring the model toward biogeochemically plausible behavior on unfamiliar soil conditions — though cross-site generalization remains challenging, with negative R$^2$ across all seeds and $\lambda$ values on our geographically distinct held-out site.
\end{abstract}

\section{Introduction}
\label{sec:introduction}

Among all greenhouse gases, nitrous oxide (N$_2$O) is the third-most important after carbon dioxide (CO$_2$) and methane (CH$_4$) \citep{Tian2024}, and the dominant ozone-depleting substance emitted in the 21st century \citep{Ravishankara2009}. Nitrous oxide emissions have grown by an unprecedented 40\% since 1980 \citep{Tian2024}, with a global warming potential (GWP) at about 273 times that of carbon dioxide \citep{Forster2021}, and a long atmospheric lifetime of approximately 117 years \citep{Prather2026}. Recent growth in N$_2$O emissions has exceeded some of the highest projected climate scenarios \citep{Tian2020}, highlighting the need for improved predictive capability. More than 70\% of nitrous oxide emissions occur as a result of agricultural processes \citep{Tian2024}, making accurate quantification of agricultural N$_2$O flux particularly essential for greenhouse gas accounting and climate mitigation efforts. Strengthening our ability to quantify nitrous oxide emissions could yield significant downstream benefits, such as more informed policy-making, better-targeted agricultural interventions to reduce emissions, and more accurate climate projections.

However, direct measurement of N$_2$O flux emissions is very challenging \citep{Pattey2007}, in part because N$_2$O atmospheric concentrations are low enough to lie outside the detection range of many analytical techniques \citep{Rapson2014}. Static chambers remain the most common method for measuring nitrous oxide fluxes from agricultural soils \citep{deKlein2020,Rapson2014}, but chamber measurements are limited in their spatial representativeness \citep{Bastos2021,Gu2013,Rochette2010}, since individual chambers typically measure emissions over small plot-scale areas \citep{Wang2001} and may require extensive deployment to characterize heterogeneous agricultural landscapes \citep{Charteris2020}. Moreover, soil nitrous oxide emissions are inherently episodic \citep{Yamulki2000}, highly variable across both time \citep{Phillips2013} and space \citep{Parkin1987}, and subject to sudden, unpredictable impacts from environmental factors like rain \citep{Jones2011} and nitrogen fertilization \citep{Phillips2013}, making chamber estimations particularly challenging \citep{Charteris2020} and more prone to biases due to hotspots \citep{Brown2024}. Using chambers to measure N$_2$O flux also tends to be either fairly costly, as in the case of automatic chambers \citep{Mastepanov2026}, or very labor-intensive, as in the case of manual chambers \citep{Hensen2013}.

Two viable and important alternatives to directly measuring N$_2$O flux emissions are to use either AI models or process-based models (like DayCent and Cycles) to simulate or predict these emissions, based on input variables that can be more easily measured. However, both options have their own shortcomings as well. 

As discussed further in Section~\ref{sec:results-and-discussion}, process-based models generally struggle to accurately capture the complex soil biogeochemical processes that govern N$_2$O production across chemically and climatically diverse soil systems without extensive site-specific calibration \citep{Butterbach-Bahl2013,Sharma2026,Joshi2024}. Process-based models also often require extensive calibration on site-specific information and experimental observations \citep{Gabbrielli2024}, limiting their portability across locations. DayCent has been shown to struggle with estimating nitrous oxide emissions in particular \citep{DelGrosso2008}, with poor agreement between DayCent-predicted and observed soil N$_2$O emissions \citep{McClelland2021}. This is almost certainly in part due to nitrous oxide's tendency to emit in spontaneous, sporadic "hot spots" and "hot moments" \citep{Kravchenko2017,Turner2016,Wagner-Riddle2020}, given that process-based models (including DayCent) are reported to systematically under-predict emissions when actual emissions are high, i.e., during hotspots \citep{Gaillard2018}, and to struggle with the temporal dynamics of N$_2$O emissions \citep{Jarecki2008}.

The application of machine learning to N$_2$O prediction remains relatively nascent \citep{Sharma2026}, with comparatively fewer studies systematically characterizing its limitations relative to the process-based modeling literature. Classical machine learning models like Random Forest have demonstrated early promise for N$_2$O prediction \citep{Philibert2013}. However, initial findings suggest that N$_2$O predictions made by ML models may suffer from many of the same issues as those made by process-based models. Like process-based models, traditional machine learning models similarly struggle to capture the complex biogeochemical processes that drive nitrous oxide emissions \citep{Gnisia2025}, and struggle to generalize beyond their training domain \citep{Sharma2026}.

In this paper, we propose and implement a Physics-Informed Neural Network (PINN) that combines data-driven prediction with biogeochemical constraints derived from the DayCent family of process-based models. First suggested by \citet{Raissi2019}, PINNs are neural networks whose loss functions contain two components: the data loss function (typically the MSE of model predictions relative to measured data) and the physics loss function (the MSE of model predictions relative to values predicted by biogeochemical equations). In the case of N$_2$O predictions, these equations represent the real-world biogeochemical conditions that constrain nitrous oxide emissions.

\citet{Vemuri2024} previously applied this framework to N$_2$O prediction — the only such attempt known to the authors — finding that physics-informed loss terms may improve predictive accuracy. However, their physics residual relied on generic Michaelis-Menten enzyme kinetics rather than equations derived specifically for soil nitrogen biogeochemistry, their dataset comprised 2,246 observations from two geographically similar Midwestern US sites \citep{Saha2021}, and their evaluation included no leave-one-site-out validation or comparison against process-based model baselines. In this paper, we address these limitations directly: we derive our physics residual from the mechanistic equations underlying the DayCent family of process-based models, train and evaluate on 8,271 daily observations across four geographically distinct US sites, and benchmark against uncalibrated Cycles simulation as a process-based baseline. We demonstrate that our PINN substantially outperforms uncalibrated Cycles simulation in holdout validation, and find that while the physics residual hurts in-distribution performance, it consistently improves performance and reduces variability in leave-one-site-out validation — suggesting a tradeoff between in-distribution accuracy and out-of-distribution robustness.

\section{Methodology}
\label{sec:methodology}

\subsection{PINN Background}
\label{subsec:background}

The loss function of PINNs is given by:

\begin{equation}
\mathcal{L} = \mathcal{L}_{data} + \lambda \mathcal{L}_{physics}
\label{eq:total-loss}
\end{equation}

\begin{center}
  \begin{tabular}{llll}
    \toprule
    Variable      & Description                & Units                 & Source   \\
    \midrule
    $\mathcal{L}$ & total loss & g$^2$ N ha$^{-2}$ d$^{-2}$ & \\
    $\mathcal{L}_{data}$ & data loss & g$^2$ N ha$^{-2}$ d$^{-2}$ & Equation~\ref{eq:data-loss} \\
    $\lambda$ & weight of the physics loss residual & unitless & hyperparameter \\
                & (relative to data loss) &         &   \\
    $\mathcal{L}_{physics}$ & physics loss residual & g$^2$ N ha$^{-2}$ d$^{-2}$ & Equation~\ref{eq:physics-loss} \\
    \bottomrule
  \end{tabular}
\end{center}

Our model's data loss is calculated as the mean-squared error (MSE) of the model-predicted N$_2$O values, relative to the measured ("actual") N$_2$O values, over $n$ training samples, following \citet{Raissi2019}.

\begin{equation}
\mathcal{L}_{data} = \frac{1}{n} \sum_{i=1}^{n}\left(\phi_{{N_2O}_i} - \hat{\phi}_{{N_2O}_i}\right)^2
\label{eq:data-loss}
\end{equation}

\begin{center}
  \begin{tabular}{llll}
    \toprule
    Variable      & Description                & Units                 & Source   \\
    \midrule
    $\mathcal{L}_{data}$ & data loss & g$^2$ N ha$^{-2}$ d$^{-2}$ & \\
    $\phi_{N_2O}$ & measured N$_2$O flux & g N ha$^{-1}$ d$^{-1}$ & DSFAS data \\
    $\hat{\phi}_{N_2O}$ & model-predicted N$_2$O flux & g N ha$^{-1}$ d$^{-1}$ & model \\
    \bottomrule
  \end{tabular}
\end{center}

Similarly, our model's physics loss is calculated as the mean-squared error (MSE) of the model-predicted N$_2$O values, relative to the values predicted by our DayCent-derived equations, following \citet{Raissi2019}.

\begin{equation}
\mathcal{L}_{physics} = \frac{1}{n} \sum_{i=1}^{n}\left(\tilde{\phi}_{{N_2O}_i} - \hat{\phi}_{{N_2O}_i}\right)^2
\label{eq:physics-loss}
\end{equation}

\begin{center}
  \begin{tabular}{llll}
    \toprule
    Variable      & Description                & Units                 & Source   \\
    \midrule
    $\mathcal{L}_{physics}$ & physics loss residual & g$^2$ N ha$^{-2}$ d$^{-2}$ &        \\
    $\tilde{\phi}_{N_2O}$ & physics-predicted N$_2$O flux & g N ha$^{-1}$ d$^{-1}$ & Equation~\ref{eq:physics-predicted-n2o-flux} \\
    $\hat{\phi}_{N_2O}$ & model-predicted N$_2$O flux & g N ha$^{-1}$ d$^{-1}$ & model prediction \\
    \bottomrule
  \end{tabular}
\end{center}

\subsection{Physics Loss Residual}
\label{subsec:physics-loss}

We now lay out the foundational equations that we use to calculate the physics-predicted N$_2$O values, drawing upon three primary papers, which form the fundamental basis of the Nitrogen Trace Gas Submodel of DayCent as described in Part 2, Section 3.5 of \citet{Hartman2018}: \citet{Parton1996}, \citet{DelGrosso2000}, and \citet{Parton2001}. We additionally draw upon \citet{Hartman2019} for the equations of specific graphs when they are not provided in the three aforementioned papers. Note that DayCent is a constantly evolving process model, and that the equations our paper uses may no longer be in use in newer generations of DayCent. As the Preface of \citet{Hartman2018} states directly: "The model attempts to document multiple versions of the DayCent model that differ in the set processes they include, the way these processes are calculated, and the format of input and output files. Different versions emerge as separate research groups modify the model to meet the requirements of
their research and applications."

Following \citet{Parton2001}, the equations for nitrous oxide flux as a result of nitrification are given by:

\begin{equation}
\phi_{NO_3} = Net_{min} \cdot K_1 + K_{max} \cdot NH_4 \cdot F_n(T_s) \cdot F_n(WFPS) \cdot F_n(pH)
\label{eq:no3-flux}
\end{equation}

\begin{center}
  \begin{tabular}{llll}
    \toprule
    Variable     & Description                 & Units                 & Source   \\
    \midrule
    $\phi_{NO_3}$ & soil nitrification flux    & g N m$^{-2}$ d$^{-1}$ &        \\
    $Net_{min}$   & daily net N mineralization & g N m$^{-2}$ d$^{-1}$ & Cycles \\
    $K_1$         & fraction of $Net_{min}$ assumed to be nitrified each day & unitless & constant at 0.20 \\
    $NH_4$        & model-derived soil ammonium concentration & g N m$^{-2}$ & Cycles \\
    $K_{max}$     & maximum fraction of $NH_4^+$ nitrified & d$^{-1}$ & constant at 0.10 \\
    $F_n(T_s)$    & effect of soil temperature on nitrification & unitless & Equation~\ref{eq:fnts} \\
    $F_n(WFPS)$   & effect of water-filled pore space on nitrification & unitless & Equation~\ref{eq:fnwfps} \\
    $F_n(pH)$     & effect of soil pH on nitrification & unitless & Equation~\ref{eq:fnph} \\
    \bottomrule
  \end{tabular}
\end{center}

\begin{equation}
\phi_{N_2O,nit} = K_2 \cdot \phi_{NO_3}
\label{eq:n2o-flux-nit}
\end{equation}

\begin{center}
  \begin{tabular}{llll}
    \toprule
    Variable      & Description                & Units                 & Source   \\
    \midrule
    $\phi_{N_2O,nit}$ & N$_2$O flux from nitrification & g N m$^{-2}$ d$^{-1}$ & \\
    $K_2$   & fraction of nitrified N lost as $\phi_{N_2O,nit}$ & unitless & constant at 0.02 \\
    $\phi_{NO_3}$ & soil nitrification flux    & g N m$^{-2}$ d$^{-1}$ & Equation~\ref{eq:no3-flux} \\
    \bottomrule
  \end{tabular}
\end{center}

Following \citet{Parton1996}, the equation for total nitrogen gas fluxes (including both nitrogen and nitrous oxide) as a result of denitrification is given below. (Note that this equation also appears in \citet{DelGrosso2000}, but with no modifications, so we cite \citet{Parton1996} as the original source.)

\begin{equation}
\phi_{N,den} = \textup{min}(\phi_{max}(NO_3),\phi_{max}(CO_2)) \cdot F_d(WFPS)
\label{eq:n-flux-den}
\end{equation}

\begin{center}
  \begin{tabular}{llll}
    \toprule
    Variable      & Description                & Units                 & Source   \\
    \midrule
    $\phi_{N,den}$ & total N gas fluxes from denitrification & g N ha$^{-1}$ d$^{-1}$ &        \\
    $\phi_{max}(NO_3)$ & max $\phi_{N,den}$ for a given NO$_3$ level, assuming high CO$_2$ & g N ha$^{-1}$ d$^{-1}$ & Equation~\ref{eq:phimaxno3} \\
    $\phi_{max}(CO_2)$ & max $\phi_{N,den}$ for a given respiration rate, assuming high NO$_3$ & g N ha$^{-1}$ d$^{-1}$ & Equation~\ref{eq:phimaxco2} \\
    $F_d(WFPS)$ & effect of WFPS on denitrification & unitless & Equation~\ref{eq:fdwfps} \\
    \bottomrule
  \end{tabular}
\end{center}

Following \citet{DelGrosso2000}, the following two equations for the ratio of nitrogen gas to nitrous oxide gas are given by:

\begin{equation}
R_{N_2/N_2O} = F_r(NO_3/CO_2) \cdot F_r(WFPS)
\label{eq:n2-n2o-ratio}
\end{equation}

\begin{center}
  \begin{tabular}{llll}
    \toprule
    Variable      & Description                & Units                 & Source   \\
    \midrule
    $R_{N_2/N_2O}$ & the ratio of N$_2$ to N$_2$O gases produced from denitrification & unitless &        \\
    $F_r(NO_3/CO_2)$ & NO$_3$-to-respiration ratio factor & unitless & Equation~\ref{eq:frno3co2} \\
    $F_r(WFPS)$ & effect of water-filled pore space on $R_{N_2/N_2O}$ & unitless & Equation~\ref{eq:frwfps} \\
    \bottomrule
  \end{tabular}
\end{center}

Using this equation, we can then calculate the amount of nitrous oxide gas produced from denitrification.

\begin{equation}
\phi_{N_2O,den} = \frac {\phi_{N,den}} {1 + R_{N_2/N_2O}}
\label{eq:n2o-flux-den}
\end{equation}

\begin{center}
  \begin{tabular}{llll}
    \toprule
    Variable      & Description                & Units                 & Source   \\
    \midrule
    $\phi_{N_2O,den}$ & N$_2$O flux from denitrification & g N ha$^{-1}$ d$^{-1}$ &        \\
    $\phi_{N,den}$ & total N gas fluxes from denitrification & g N ha$^{-1}$ d$^{-1}$ & Equation~\ref{eq:n-flux-den} \\
    $R_{N_2/N_2O}$ & the ratio of N$_2$ to N$_2$O gases produced from denitrification & unitless & Equation~\ref{eq:n2-n2o-ratio}       \\
    \bottomrule
  \end{tabular}
\end{center}

Then, finally, by drawing upon all of the prior equations:

\begin{equation}
\tilde{\phi}_{{N_2O}_i} = \alpha_{m^2\to ha} \cdot \phi_{{N_2O,nit}_i} + \phi_{{N_2O,den}_i}
\label{eq:physics-predicted-n2o-flux}
\end{equation}

\begin{center}
  \begin{tabular}{llll}
    \toprule
    Variable      & Description                & Units                 & Source   \\
    \midrule
    $\tilde{\phi}_{N_2O}$ & physics-predicted N$_2$O flux & g N ha$^{-1}$ d$^{-1}$ & \\
    $\alpha_{m^2\to ha}$ & units conversion factor from m$^{-2}$ to ha$^{-1}$ & m$^2$ ha$^{-1}$ & constant at $10^4$ \\
    $\phi_{N_2O,nit}$ & N$_2$O flux from nitrification & g N m$^{-2}$ d$^{-1}$ & Equation~\ref{eq:n2o-flux-nit}       \\
    $\phi_{N_2O,den}$ & N$_2$O flux from denitrification & g N ha$^{-1}$ d$^{-1}$ & Equation~\ref{eq:n2o-flux-den}       \\
    \bottomrule
  \end{tabular}
\end{center}

The other equations we used, as well as their origins in the research literature, can be found in Section~\ref{subsec:auxiliary-equations}.

\subsection{Dataset}
\label{subsec:dataset}

All of the variables that we use in our work can be categorized into one of three categories:

\begin{enumerate}
    \item "Input Variables": The "starting" variables whose data is either directly measured or directly simulated by Cycles. These are also the same predictor variables that are fed directly into the model as input features.
    \item "Intermediate Variables": The variables that are derived from the input variables using the equations described in Section~\ref{subsec:physics-loss} and Section~\ref{subsec:auxiliary-equations}
    \item "Final Variables": The variables that are directly used to calculate the model loss and evaluate its performance; namely: $\phi_{N_2O}$ (measured N$_2$O flux), $\hat{\phi}_{N_2O}$ (model-predicted N$_2$O flux), and $\tilde{\phi}_{N_2O}$ (physics-predicted N$_2$O flux).
\end{enumerate}

The following table is an exhaustive list of the 10 input variables we used.

\begin{table}[H]
  \caption{Input Variables}
  \label{table:input-variables}
  \centering
  \begin{tabular}{llll}
    \toprule
    Variable      & Description                & Units                 & Source   \\
    \midrule
    $NO_3$ & soil nitrate concentration & mg N kg$^{-1}$ & Cycles \\
    $NH_4$ & model-derived soil ammonium concentration & g N m$^{-2}$ & Cycles \\
    $pH$ & soil pH in the 2nd soil layer & unitless & static site measurement \\
    $T_s$ & average soil temperature of soil layers 2 and 3 & $^{\circ}$C & Cycles \\
    $T_{a,max}$ & maximum monthly air temperature & $^{\circ}$C & Cycles-approximated \\
    $\%S$ & percentage of soil that is sand & unitless & site survey \\
    $\%C$ & percentage of soil that is clay & unitless & site survey \\
    $Net_{min}$ & daily net N mineralization & g N m$^{-2}$ d$^{-1}$ & Cycles \\
    $VWC_l$ & volumetric water content of soil layer $l$ & m$^3$ m$^{-3}$ & Cycles \\
    $BD_l$ & soil bulk density of soil layer $l$ & g cm$^{-3}$ & Cycles \\
    \bottomrule
  \end{tabular}
\end{table}

$\phi_{N_2O}$, our measured N$_2$O flux, is taken from \citet{Robertson2026}, an online living dataset hosted by the KBS LTER (Kellogg Biological Station Long-Term Ecological Research) program at Michigan State University, and part of the USDA-sponsored Data Science for Food and Agricultural Systems (DSFAS) project. We accessed the data on June 19, 2026; the dataset has likely been updated and modified since then.

Our dataset consists of data drawn from four sites (some of which were treated with various treatment types), enumerated as follows:

\begin{enumerate}
    \item An agricultural field near Ames, IA
    \item KBS GLBRC, Hickory Corners, MI, USA (Treatments G1, G2, G3, G4)
    \item Kelly North, Iowa State Research Farm, Ames, Iowa
    \item Cook Agronomy Farm, No-till (AmeriFlux US-RC1) (Treatments NT and CT)
\end{enumerate}

With the assistance of \citeauthor{Robertson2026} project collaborators, we traced through tillage history to characterize Sites 1 and 3 as conventionally tilled, and Site 2 as untilled. Site 4 with treatment "CT" was conventionally tilled, and Site 4 with treatment "NT" was not tilled. For Equation~\ref{eq:frwfps} as in \citet{DelGrosso2000}, we equate "conventionally tilled" with "repacked", and "not tilled" as "intact".

In cases where our equations expect site-specific values, but our data only provides layer-specific values, we take the weighted average across layers 2 and 3 (under \citet{Hartman2019}'s layer convention), where the weight for each layer is the layer thickness. This is the same approach we use in, for instance, Equation~\ref{eq:wcrel}.

Our measured N$_2$O flux data was preprocessed as follows: Sub-daily measurements were aggregated to daily values to match the daily time step of the Cycles model. Flux measurements were then averaged across field replications, as Cycles does not simulate replication. For sites with measurements at multiple landscape elevations, only data from the highest elevation were retained to maintain consistency with the one-dimensional structure of the Cycles model. This preprocessing was performed prior to delivery of the dataset used in this work.

Our data on the sand and clay percentages (\%S and \%C, respectively) was sourced from gSSURGO/SoilGrids or original site publications. In our variable tables, we hereafter simply list the source as ``site survey".

With the sole exception of soil pH (which was a static site-level measurement), all of our input variables are directly simulated by Cycles. Although measured data on most of the input variables we need was available via the DSFAS project dataset, the measured data had serious gaps. Some sites lacked measurements of certain key variables like net mineralization and soil bulk density, and even when present, the measurements were often taken at varying depths. Rather than taking a "mix-and-match" approach of combining measured data taken from the DSFAS dataset with simulated data taken from Cycles, we instead opted to go with the more cohesive, consistent approach of using entirely Cycles-simulated data.

Our reasoning is as follows: Our DayCent-derived equation chain maps a joint soil state (NO$_3$, NH$_4$, T$_S$, VWC, BD, \%S, \%C…) to a single predicted N$_2$O value. This residual is only physically meaningful if the input vector at each timestep is a jointly realizable state, i.e., something that could actually co-occur in one soil at one time. Our approach of using entirely Cycles-simulated data guarantees this condition, because it generated every driver under one internally-coupled dynamics. For instance, Cycles-simulated NO$_3$ and VWC data at day $t$ are produced by the exact same simulated moisture/temperature trajectory. Gap-filling (the alternative mix-and-match approach) would break this condition: By filling a single variable at time $t$ with both Cycles-simulated data and measured data, we'd be feeding the equations a state vector that no coherent soil ever occupied. 

Moreover, filling gaps in measured data with Cycles-simulated data would render feature provenance (i.e., data sourcing) as a function of site. More specifically: Measured data from the DSFAS dataset is available for some sites but not for other sites, so if our model were to struggle in leave-one-site-out validation, it'd be impossible to distinguish between the model's struggle to generalize geographically and the model's struggle to generalize between simulated and measured data. For instance, suppose our model is trained on Site A, which has mostly measured data, and then tested on Site B, which has mostly Cycles-simulated data. The model's struggle to generalize across geographic sites would appear externally identical to its struggle to generalize between the measured data of Site A and the simulated data of Site B. In summary: augmenting the DSFAS measured data with simulated data would confound provenance with geographic site identity.

\subsection{Model Architecture \& Training}
\label{subsec:model-architecture-and-training}

We use PyTorch to implement a three-layer MLP ($10 \to 64 \to 32 \to 1$), with ReLU activation and dropout (p=0.3) applied after each of the first two linear layers. Our model utilizes the Adaptive Moment Estimation (Adam) optimizer with learning rate 0.001. As mentioned in Section~\ref{subsec:background}, we use MSE for all of our loss functions. Our model also uses early stopping with patience=20 to mitigate overfitting on training data. Our findings as discussed in Section~\ref{sec:results-and-discussion} were calculated as the mean across ten random seeds (41 through 50) unless explicitly stated otherwise. 

We make use of two distinct model evaluation techniques: holdout validation, and leave-one-site-out validation. In the case of holdout validation, we use a random 80/20 training/testing data split, and then further divide the training data using an 80/20 training/validation data split. The training data is the data that the model is actually trained on, while the validation dataset is used to evaluate model performance for the early stopping mechanism. This is necessary because using the testing dataset directly for early stopping would be a form of data leakage. In the case of leave-one-site-out validation, we use data from Site 4 as our testing data, since Site 4 is most distinct from the rest of our dataset, both geographically and in terms of soil characteristics (i.e., pH, soil texture, etc., as discussed further in Section~\ref{sec:results-and-discussion}), and thus, the most demanding test of geographic generalizability. Sites 1–3 are split 80/20 into training and validation sets, with the validation set used exclusively for early stopping.

\section{Results \& Discussion}
\label{sec:results-and-discussion}

\subsection{Model Performance Relative to Cycles}
\label{subsec:model-performance}

Unless otherwise noted, per-site results in this subsection are reported for seed 41 and $\lambda=0$; aggregate results across all seeds are provided in Section~\ref{subsec:physics-residual-impact} and Appendix~\ref{subsec:additional-tables}.

Our baseline model ($\lambda=0$) outperformed Cycles dramatically for every site, achieving overall R$^2=0.423$ on seed 41 against Cycles' mean of R$^2=0.01$, with even the worst-performing configuration ($\lambda=1$, mean R$^2=0.090$) outperforming Cycles by nearly an order of magnitude. We also find, perhaps unsurprisingly, that our model performed best on Site 2, for which it had the most training data by a wide margin. By comparison, its test R$^2$ value for Sites 1 and 4 were worse by more than 0.15.

\begin{table}[h]
  \caption{Per-site R$^2$ summary (holdout validation, seed=41, $\lambda$=0)}
  \label{table:per_site}
  \centering
  \begin{tabular}{llrrrr}
    \toprule
    Site & Dataset Size & Split & Model R$^2$ & Cycles R$^2$ \\
    \midrule
    1 & 435 & Train & 0.228 & -0.244 \\
      &     & Test  & 0.264 & -0.407 \\
    2 & 6,468 & Train & 0.490 & 0.122 \\
      &       & Test  & 0.490 & 0.181 \\
    3 & 224 & Train & 0.645 & -0.376 \\
      &     & Test  & -0.299 & -0.829 \\
    4 & 1,144 & Train & 0.329 & -0.210 \\
      &       & Test  & 0.333 & -0.259 \\
    \midrule
    Overall & 8,271 & Train & 0.420 & 0.011 \\
            &       & Test  & 0.442 & 0.003 \\
            &       & All   & 0.423 & 0.010 \\
    \bottomrule
  \end{tabular}
\end{table}

Our model scores a remarkably low R$^2=-0.299$ on Site 3, but this can be 
explained by the occurrence of a single outlier. Near late July of 2006 in 
our data, our model predicted a spike of nearly 100 g N ha$^{-1}$ 
d$^{-1}$, more than three times the measured flux of approximately 
30 g N ha$^{-1}$ d$^{-1}$. Because R$^2$ penalizes errors in proportion 
to their squared magnitude, a single sufficiently large prediction error 
can dominate the metric. When this outlier is removed, Site 3 test R$^2$ 
jumps from $-0.299$ to $0.518$, consistent with performance at our other 
sites.

For further analysis, we now turn our attention to the two sites that both our model and Cycles performed best and worst at, respectively: Site 2 and Site 1. 

\begin{figure}
    \centering
    \includegraphics[width=0.9\linewidth]{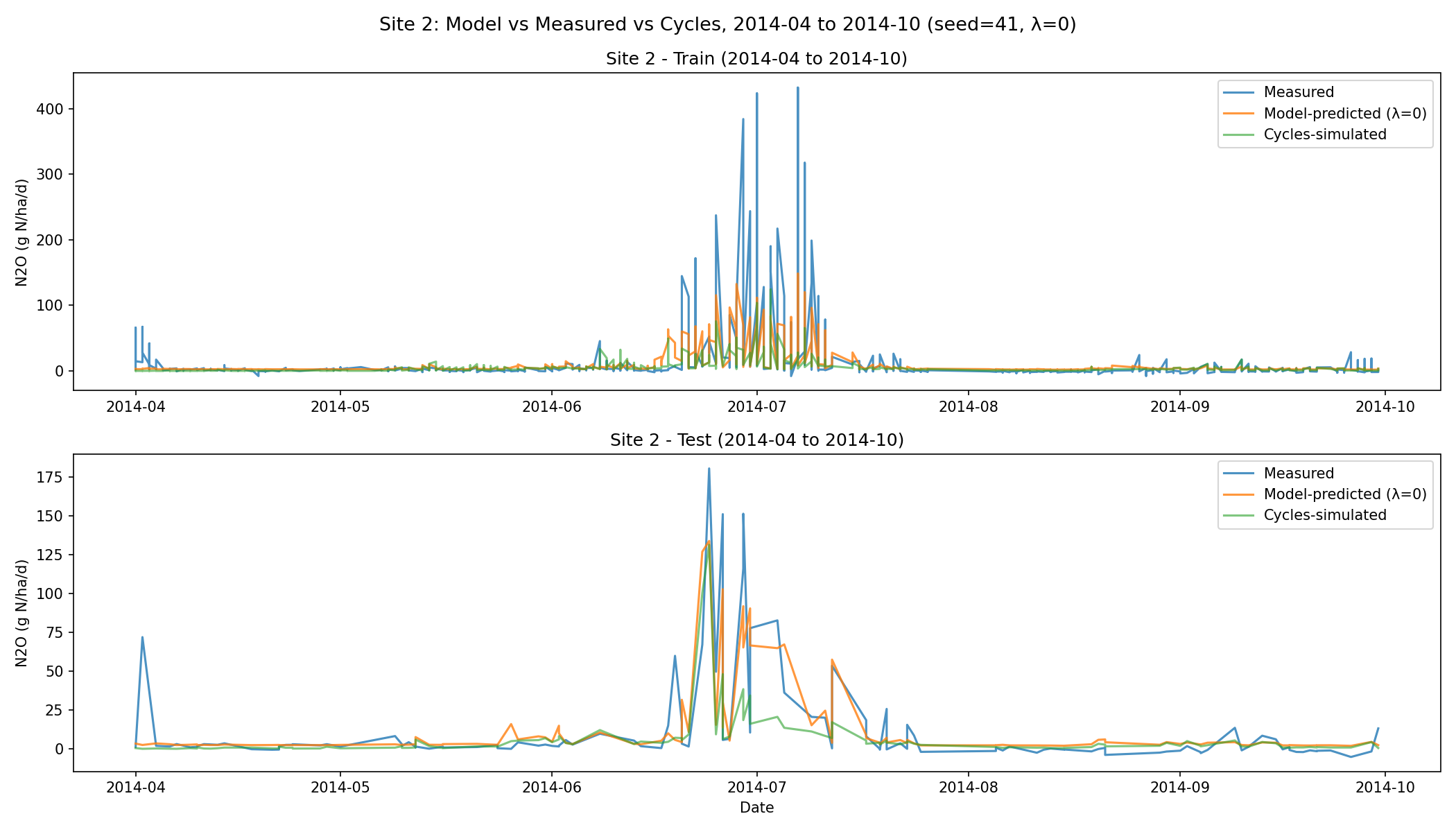}
    \caption{Site 2 time series, April--October 2014 (most active 6-month emission period)}
    \label{fig:site-2-zoomed}
\end{figure}

Figure~\ref{fig:site-2-zoomed} illustrates N$_2$O dynamics at Site 2 during the most active emission period (April--October 2014). Both our model and Cycles struggled with N$_2$O's characteristic pattern of near-zero background emissions punctuated by sudden, volatile hotspots. Cycles consistently underestimated peak emissions by almost a full order of magnitude. Our model generally matched the timing of emission events more closely, though it still somewhat underestimated peak magnitudes.

At Site 1, both our model and Cycles failed to capture most N$_2$O hotspots (see Figure~\ref{fig:site-1} in the Appendix). Our model overestimated background emissions relative to Cycles, but neither approach responded adequately to peak emission events. We speculate this reflects underrepresentation in training data, as Site 1 comprised only $\frac{435}{8271} = 5.26\%$ of the total dataset. (See Figures~\ref{fig:site-2}--\ref{fig:site-4} in the Appendix for the complete time series.)

\subsection{Physics Residual: In-Distribution vs. Out-of-Distribution Performance}
\label{subsec:physics-residual-impact}

\begin{figure}[h]
    \centering
    \begin{minipage}{0.48\linewidth}
        \centering
        \includegraphics[width=\linewidth]{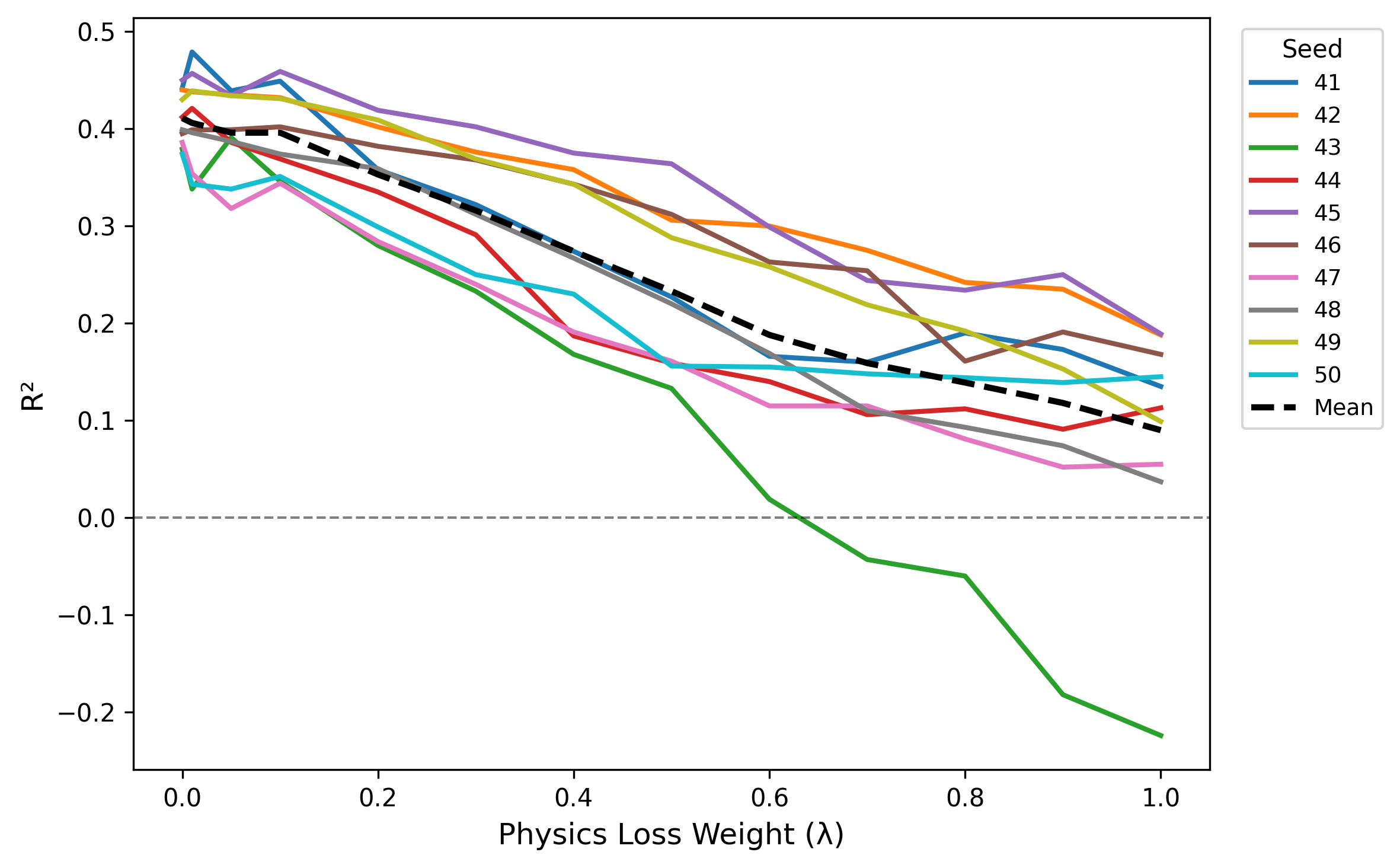}
        \caption{R$^2$ across $\lambda$ sweep (holdout validation). Individual seed trajectories (seeds 41--50). Physics constraints consistently degrade in-distribution performance, with degradation increasing at higher $\lambda$. Seed 43 exhibits notably high sensitivity at large $\lambda$ values.}
        \label{fig:lambda-sweep-holdout}
    \end{minipage}
    \hfill
    \begin{minipage}{0.48\linewidth}
        \centering
        \includegraphics[width=\linewidth]{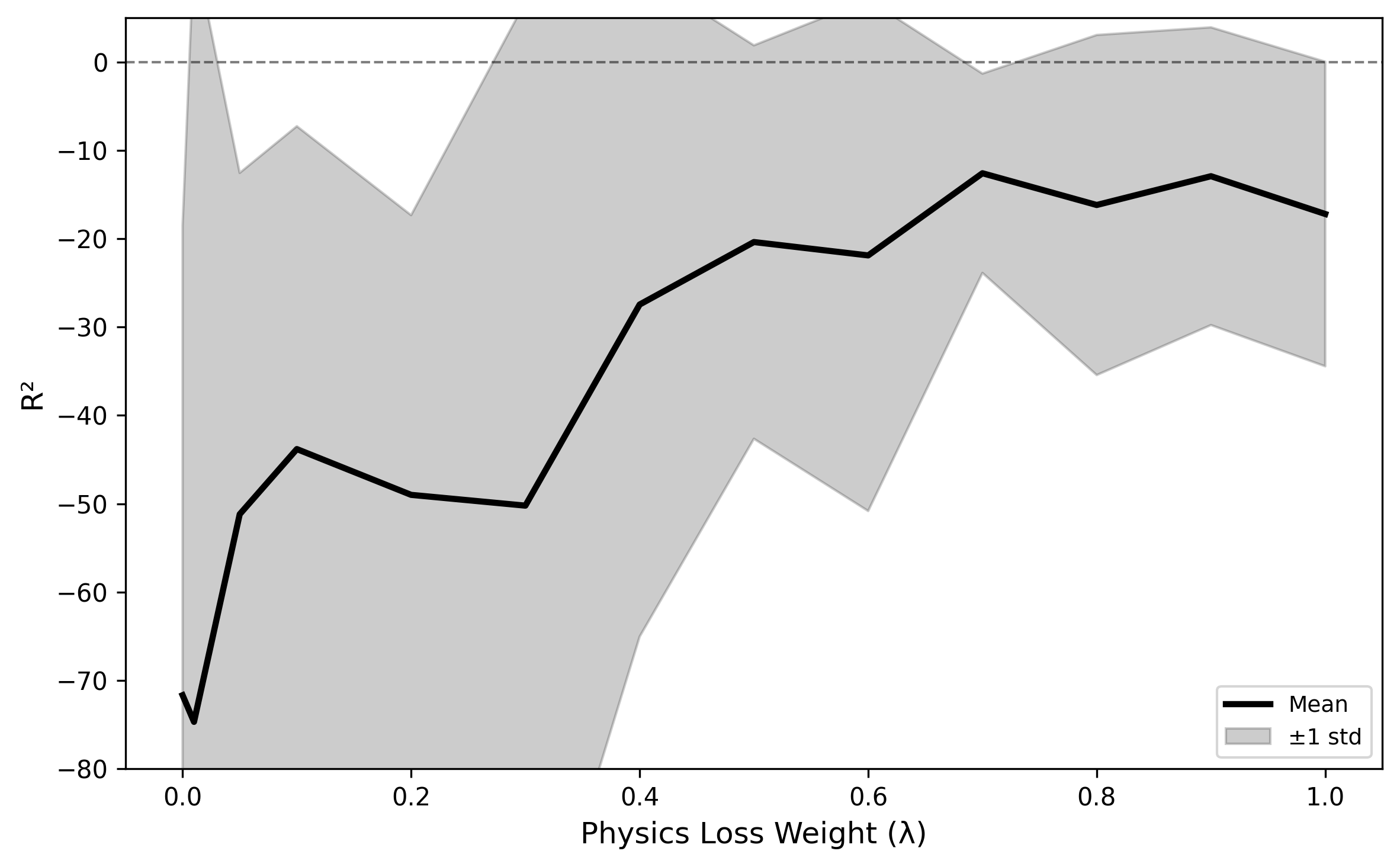}
        \caption{R$^2$ across $\lambda$ sweep (leave-one-site-out validation). Mean $\pm$1 std across seeds 41--50. Physics constraints consistently improve out-of-distribution performance, peaking at $\lambda=0.7$. Values below $-80$ are clipped; see Table~\ref{table:lambda-sweep-loso} in the Appendix for complete values.}
        \label{fig:lambda-sweep-loso}
    \end{minipage}
\end{figure}

From Figure~\ref{fig:lambda-sweep-holdout}, we see that in the case of holdout validation, the physics constraints imposed by our physics loss residual $\mathcal{L}_{physics}$ consistently hurt model performance across all tested values of $\lambda$, with model performance degradation increasing as physics weight $\lambda$ increases. We additionally see that, as $\lambda$ increases, the mean standard deviation across the ten seeds increases, suggesting that a higher physics weight results in greater performance variability. (See Table~\ref{table:lambda-sweep-holdout} in the Appendix for complete results.)

This general pattern is flipped in the case of leave-one-site-out validation. From Figure~\ref{fig:lambda-sweep-loso}, we see that although our model's performance is severely negative, with negative R$^2$ values across all seeds and $\lambda$ values, the general trend is the opposite of what we saw in holdout validation. As $\lambda$ increases, mean R$^2$ increases and mean standard deviation decreases, suggesting that a higher physics weight results in \textit{improved} model performance and \textit{less} performance variability. Our model's performance peaks at $\lambda=0.7$, where mean R$^2=-12.597$, a substantial improvement from the MLP baseline $\lambda=0.00$, where mean R$^2=-71.672$. Moreover, the mean standard deviation at $\lambda=0.7$ reaches its lowest value of $11.281$, insufficient to explain the significant gap in performance between $\lambda=0.7$ and $\lambda=0.0$ means. (See Table~\ref{table:lambda-sweep-loso} in the Appendix for complete results.)

To explain this apparent contradiction, we turn to recent findings from \citet{DelGrosso2026} discussing \citet{Sharma2026}: that process-based models accurately represent the drivers of N$_2$O emissions (soil water content, mineral N, temperature) but systematically struggle to predict the fraction of cycled N (produced from denitrification) that actually gets emitted as N$_2$O. This is because the nitrification and denitrification processes that govern N$_2$O production are sufficiently complex that current equations necessarily rely on simplifying assumptions. For instance, the use of water-filled pore space (WFPS) as a factor in N$_2$O production is generally considered non-ideal at best \citep{Farquharson2008}, but functions based on water filled pore space are still widely used (such as in DayCent), in the absence of widely validated alternatives.

Beyond conceptual limitations like WFPS, our physics residual follows well-established research literature in using fixed biogeochemical constants like $K_1=0.20, K_2=0.02, K_{max}=0.10$, etc. Ultimately, these constants are widely accepted but nonetheless flawed simplifications that hinder our physics residual's ability to accurately model real-world physical constraints. In summary, even a perfectly implemented physics residual with perfectly derived equations would be expected to provide weak or inconsistent regularization signal on heterogeneous multi-site data, because the biogeochemical constants themselves are the source of uncertainty. This simultaneously explains the poor performance of both the uncalibrated Cycles model and our PINN's physics residual.

Consequently, we suspect that the physics residual hurts in-distribution (i.e., in holdout validation) because, as suggested by \citet{DelGrosso2026}, fixed constants are too coarse to improve local accuracy, and the data alone is a much better predictor of N$_2$O emissions.

However, the physics residual helps out-of-distribution (i.e., in leave-one-site-out validation) because even imperfect physics constraints are better than nothing when the model encounters unfamiliar soil conditions. More precisely: the same biogeochemical constants that are too imprecise to improve model performance in holdout validation, are simultaneously precise enough to prevent the model from making biogeochemically implausible predictions on unseen soil types. In the case of our leave-one-site-out validation testing, the soil characteristics of our left-out test site, Site 4, are substantially different from those of our other 3 training sites. Site 4 had an average soil pH of 5.2 and average soil texture of 21.0\% clay and 11.3\% sand, where the other 3 sites had an average soil pH of 7.5 and average soil texture of 12.3\% clay and 61.4\% sand. In the face of such unfamiliar soil conditions, our physics residual may serve to help anchor the model to biogeochemically plausible behavior.

If this reasoning is correct, then we conclude that, while the physics residuals of PINNs are not magic performance boosters, they may serve as an important fallback: when the model needs to generalize across geographically distinct sites, the physics residual offers the tradeoff of sacrificing in-distribution accuracy for out-of-distribution robustness.

\section{Conclusion}
\label{sec:conclusion}

In summary, our model achieves mean R$^2$ = 0.411 in holdout validation, dramatically outperforming Cycles (mean R$^2$ = 0.010), but struggles significantly in leave-one-site-out validation, achieving negative R$^2$ at every value of $\lambda$ across all ten seeds, suggesting that geographic generalizability remains challenging. Our model's best performance in holdout validation occurred at $\lambda \approx 0$, while its best performance in leave-one-site-out validation occurred at $\lambda \approx 0.7$ --- an approximately 6× improvement in mean R$^2$ over the MLP baseline --- suggesting that physics constraints sacrifice in-distribution accuracy for out-of-distribution robustness, anchoring the model toward biogeochemically plausible behavior on unfamiliar soil conditions.

Our paper does have notable methodological weaknesses. Firstly, with the sole exception of pH and measured N$_2$O, all of our data was Cycles-simulated. Cycles-derived inputs carry whatever systematic biases or errors Cycles has, meaning that from the very start, our PINN's predictive accuracy was bounded by how well Cycles represents the true soil state. Secondly, our paper does not make use of collocation points as \citet{Raissi2019} did in their original PINN. Our setting does not naturally lend itself to collocation points given its discrete daily timestep structure, meaning our physics loss is enforced only at observed data points, which may limit the regularization effect of the physics constraint. Thirdly, our physics residual relies on fixed biogeochemical constants (e.g., $K_1=0.20$, $K_2=0.02$, $K_{max}=0.10$) that are widely accepted simplifications but do not vary across sites or soil conditions, limiting the residual's ability to provide meaningful regularization signal on heterogeneous multi-site data. Fourthly, although our data is sourced from Cycles, our equations are derived from the DayCent family; this mismatch may result in the physics residual's performance being hampered by conflating the internal biases of two different process-based models. Finally, our dataset was not evenly geographically distributed --- our assessment of the model's overall performance was heavily dominated by its performance at Site 2, due to Site 2's overrepresentation in the data (constituting 6,468 of 8,271 points, or 78.2\% of our entire dataset).

Future work should experiment with different model architectures beyond MLPs (such as PINNsformers as in \citet{Zhao2024} or S-Pformers as in \citet{Arni2025}), as well as different process-based models and physics residuals (such as implementing Cycles equations for a more direct comparison with uncalibrated Cycles performance). We furthermore believe that more extensive leave-one-site-out experimentation, particularly with larger and more geographically diverse datasets, could help validate our hypothesis that physics residuals improve out-of-distribution robustness.

\begin{ack}

F.Y. was supported by the 2026 Student Mentoring and Research Training (SMaRT) program provided by The Science Alliance, which is a Tennessee Higher Education Commission (THEC) center of excellence administered by The University of Tennessee-Oak Ridge Innovation Institute (UT-ORII).

This work was supported in part by USDA NIFA, Award \#2023-67021-40008.

We thank the DSFAS project team at KBS LTER for providing the N$_2$O flux dataset used in this work.

\end{ack}

\bibliographystyle{plainnat}
\bibliography{references}


\appendix

\section{Technical appendices and supplementary material}
\label{sec:appendix}

\subsection{Auxiliary Equations}
\label{subsec:auxiliary-equations}

Following \citet{Hartman2019}, the equations for the respective effects of soil temperature, water-filled pore space, and soil pH on nitrification are given below. $F_n(T_s)$ is computed via a generalized Poisson density function following \citet{Hartman2019} (Equations 3.13--3.14), parameterized by the long-term maximum monthly air temperature ($T_{a,max}$).

\begin{equation}
F_n(T_s) = 
\begin{cases}
\left(\frac{A_1 - T_s}{A_1 - T_{a,max}}\right)^{A_2} \cdot \exp\left(\frac{A_2}{A_3}\left(1.0 - \left(\frac{A_1 - T_s}{A_1 - T_{a,max}}\right)^{A_3}\right)\right) & \text{if } T_{a,max} \geq 35.0 \\[10pt]
f\_gen\_poisson\_density\left(T_s + (A_0 - T_{a,max}),\ A_0,\ A_1,\ A_2,\ A_3\right) & \text{if } T_{a,max} < 35.0
\end{cases}
\label{eq:fnts}
\end{equation}

\begin{center}
  \begin{tabular}{llll}
    \toprule
    Variable     & Description                 & Units                 & Source   \\
    \midrule
    $F_n(T_s)$ & effect of soil temperature on nitrification & unitless & \\
    $T_s$ & average soil temperature of soil layers 2 and 3 & $^{\circ}$C & Cycles \\
     $A_0$ & peak location parameter (bell curve center) & $^{\circ}$C & constant at 35.0 \\
    $A_1$ & lower bound parameter & $^{\circ}$C & constant at -5.0 \\
    $A_2$ & spread shape parameter & unitless & constant at 4.5 \\
    $A_3$ & asymmetry shape parameter & unitless & constant at 7.0 \\
    \bottomrule
  \end{tabular}
\end{center}

\citet{Hartman2019} refers to $T_{a,max}$ somewhat ambiguously as the "long-term maximum monthly air temperature", which could be interpreted to mean that $T_{a,max}$ is a site-specific value, calculated by taking the average temperature across each month for a site, then finding the maximum across all of them. However, \citet{Parton2001} states on Page 17,406: "The effect of soil temperature on nitrification is based on data presented by \citet{Malhi1981} which show that there is an optimal temperature for nitrification that is a function of the average maximum monthly air temperature for the warmest month of the year." This statement could also be reasonably interpreted to mean that the "average maximum monthly air temperature for the warmest month of the year" refers to $T_{a,max}$, such that $T_{a,max}$ is calculated by finding the warmest month of the year, extracting that month's average temperature.

In summary, there are essentially three possible different interpretations of how to define $T_{a,max}$, each of which are defensible in their own way: (1) maximum of monthly mean of average daily temperature $T_{avg}$, (2) maximum of monthly mean of maximum daily temperature $T_{max}$, and (3) mean of daily maximum temperature $T_{max}$ in the warmest month.

In the face of these two ambiguous and somewhat conflicting variable descriptions from \citet{Hartman2019} and \citet{Parton2001}, we choose to define "maximum monthly air temperature" as the monthly mean maximum temperature, and move forward with Interpretation 2. This is supported by Section 3.6.1 of the DayCent manual, which defines the variable TMX2M (1-12) $^{\circ}$C as the "mean monthly maximum air temperature" \citep{Hartman2018}. We therefore define $T_{a,max}$ as the maximum of TMX2M across all months.

Although neither \citet{Hartman2019} nor \citet{Parton2001} explicitly offer a formula for $T_{a,max}$, our interpretation as described above allows us to establish the following equation as our definition of $T_{a,max}$.


\begin{equation}
T_{a,\max} = \max_{M \in \{1,\ldots,12\}} \left( T_{a,M} \right)
\label{eq:tamax}
\end{equation}

\begin{center}
  \begin{tabular}{llll}
    \toprule
    Variable     & Description                 & Units                 & Source   \\
    \midrule
    $T_{a,max}$ & long-term maximum monthly air temperature & $^{\circ}$C & \\
    $T_{a,M}$ & monthly mean maximum temperature & $^{\circ}$C & Cycles \\
    \bottomrule
  \end{tabular}
\end{center}

$F_n(WFPS)$ is computed via 

\begin{equation}
F_n(WFPS) = 
\begin{cases}
\dfrac{1.0}{1.0 + 30.0 \cdot \exp(-9.0 \cdot wc_{rel})} & \text{if } wc_{rel} \leq 1.0 \\[10pt]
\dfrac{- 1.0}{1.0 - \theta_{fc}} \cdot (WFPS - 1.0) & \text{if } wc_{rel} > 1.0
\end{cases}
\label{eq:fnwfps}
\end{equation}

\begin{center}
  \begin{tabular}{llll}
    \toprule
    Variable     & Description                 & Units                 & Source   \\
    \midrule
    $F_n(WFPS)$ & effect of water-filled pore space on nitrification & unitless & \\
    $wc_{rel}$ & weighted average relative water content in the 2nd and 3rd soil layers & unitless & Equation~\ref{eq:wcrel} \\ 
    $\theta_{fc}$ & volumetric water content at field capacity & unitless & Equation~\ref{eq:theta_fc} \\
    $WFPS$ & weighted average water-filled pore space in 2nd and 3rd soil layers & unitless & Equation~\ref{eq:wfps} \\
    \bottomrule
  \end{tabular}
\end{center}

The upcoming subequations follow from \citet{Hartman2019}, under the definition of $wc_{rel}$ used herein, where $\theta_{fc,l}$ denotes $\theta_{fc}$ evaluated for layer $l$ using that layer's sand and clay fractions. Note that $width_l$ can be found in the "soilLayersCN.txt" Cycles output file, but is technically a "Cycles configuration" (as opposed to a typical Cycles-simulated value) since there is no universal Cycles convention for layer widths and labeling. Correspondingly, note that \citet{Hartman2019} did not provide an explicit convention for "2nd and 3rd soil layers", so we choose to let the 2nd and 3rd layers refer to the depths of 5 cm to 10 cm, and 10 to 20 cm, below the surface, respectively.

\begin{subequations}
\begin{align}
wc_{rel} &= \frac{\sum_{l=2}^{3} width_l \cdot wc_{rel,l}}{\sum_{l=2}^{3} width_l} \label{eq:wcrel} \\
wc_{rel,l} &= \frac{\dfrac{swc_l}{width_l} - swclimit_l}{\theta_{fc,l} - swclimit_l} \label{eq:wcrel_lyr}
\end{align}
\end{subequations}

\begin{center}
  \begin{tabular}{llll}
    \toprule
    Variable     & Description                 & Units                 & Source   \\
    \midrule
    $wc_{rel}$ & weighted average relative water content in the 2nd and 3rd soil layers & unitless & \\ 
    $wc_{rel,l}$ & relative water content of soil layer $l$ & unitless & \\
    $l$ & specific soil layer index & unitless & constant at 2 or 3 \\
    $width_l$ & thickness of soil layer $l$ & cm & Cycles \\
    $swc_l$ & soil water content of soil layer $l$ & cm H$_2$O & Equation~\ref{eq:swcl} \\
    $swclimit_l$ & minimum volumetric soil water content (wilting point) of layer $l$ & unitless & Equation~\ref{eq:swclimit} \\
    $\theta_{fc,l}$ & volumetric soil water content of layer $l$ at field capacity & unitless & Equation~\ref{eq:theta_fc} \\
    \bottomrule
  \end{tabular}
\end{center}

Since Cycles only provides volumetric water content, we derive $swc_l$ via the standard conversion formula from volumetric water content to equivalent water depth:

\begin{equation}
swc_l = VWC_l \cdot width_l
\label{eq:swcl}
\end{equation}

\begin{center}
  \begin{tabular}{llll}
    \toprule
    Variable     & Description                 & Units                 & Source   \\
    \midrule
    $swc_l$ & soil water content of soil layer $l$ & cm H$_2$O & \\
    $VWC_l$ & volumetric water content of soil layer $l$ & m$^3$ m$^{-3}$ & Cycles \\
    $width_l$ & thickness of soil layer $l$ & cm & Cycles \\
    \bottomrule
  \end{tabular}
\end{center}

We derive $swclimit_l$ from Equation 2 of \citet{Saxton1986}'s pedotransfer functions, for when the applied water tension reaches 1500 kPa; resulting in the equation $\Psi = 1500 = A_{sx} swclimit_l ^{B_{sx}}$, which can be rearranged into the form

\begin{equation}
swclimit_l = \left(\dfrac{1500}{A_{sx}}\right)^{\frac{1}{B_{sx}}}
\label{eq:swclimit}
\end{equation}

\begin{center}
  \begin{tabular}{llll}
    \toprule
    Variable     & Description                 & Units                 & Source   \\
    \midrule
    $swclimit_l$ & minimum volumetric soil water content (wilting point) of layer $l$ & unitless & \\
    $A_{sx}$ & air entry potential scaling parameter & kPa & Equation~\ref{eq:A_sx} \\
    $B_{sx}$ & pore size distribution shape parameter & unitless & Equation~\ref{eq:B_sx} \\
    \bottomrule
  \end{tabular}
\end{center}

Similarly, the following two subequations come from \citet{Hartman2019} and from the definition of $WFPS$ used herein. Note that $PD$'s value is an assumption from Equation~\ref{eq:pc}.

\begin{subequations}
\begin{align}
WFPS &= \frac{\sum_{l=2}^{3} width_l \cdot WFPS_l}{\sum_{l=2}^{3} width_l} \label{eq:wfps} \\
WFPS_l &= \frac{\dfrac{swc_l}{width_l}}{1.0 - \dfrac{BD_l}{PD}} \label{eq:wfps_lyr}
\end{align}
\end{subequations}

\begin{center}
  \begin{tabular}{llll}
    \toprule
    Variable     & Description                 & Units                 & Source   \\
    \midrule
    $WFPS$ & weighted average water-filled pore space in 2nd and 3rd soil layers & unitless & \\
    $WFPS_l$ & water-filled pore space of specific soil layer $l$ & unitless & \\
    $l$ & specific soil layer index & unitless & constant at 2 or 3 \\
    $width_l$ & thickness of the layer & cm & Cycles \\
    $swc_l$ & soil water content of soil layer $l$ & cm H$_2$O & Equation~\ref{eq:swcl} \\
    $BD_l$ & soil bulk density of soil layer $l$ & g cm$^{-3}$ & Cycles \\
    $PD$ & particle density & g cm$^{-3}$ & constant at 2.65 \\ 
    \bottomrule
  \end{tabular}
\end{center}

$F_n(pH)$ is computed via 

\begin{equation}
F_n(pH) = B_1 + \frac{B_2}{\pi} \cdot \arctan\left(\pi \cdot B_3 \cdot (pH - B_0)\right)
\label{eq:fnph}
\end{equation}

\begin{center}
  \begin{tabular}{llll}
    \toprule
    Variable     & Description                 & Units                 & Source   \\
    \midrule
    $F_n(pH)$     & effect of soil pH on nitrification & unitless & \\
    $pH$ & soil pH in the 2nd soil layer & unitless & static site measurement \\
    $B_0$ & inflection point parameter & unitless & constant at 5.0 \\
    $B_1$ & vertical offset parameter & unitless & constant at 0.56 \\
    $B_2$ & amplitude parameter & unitless & constant at 1.0 \\
    $B_3$ & sigmoid transition steepness parameter & unitless & constant at 0.45 \\
    \bottomrule
  \end{tabular}
\end{center}

The following equation comes from \citet{Parton1996}. Note that we calculate $NO_3$ as the weighted average across layers 2 and 3, as per \citet{Hartman2019}'s convention (where the weights are the soil layer widths).

\begin{equation}
\phi_{max}(NO_3) = 11,000 + \frac{40,000 \cdot \arctan(\pi \cdot 0.002 \cdot (NO_3 - 180))}{\pi}
\label{eq:phimaxno3}
\end{equation}

\begin{center}
  \begin{tabular}{llll}
    \toprule
    Variable      & Description                & Units                 & Source   \\
    \midrule
    $\phi_{max}(NO_3)$ & max $\phi_{N,den}$ for a given NO$_3$ level, assuming high CO$_2$ & g N ha$^{-1}$ d$^{-1}$ & \\
    $NO_3$ & soil NO$_3$ concentration & mg N kg$^{-1}$ & Cycles \\
    \bottomrule
  \end{tabular}
\end{center}

\begin{equation}
\phi_{max}(CO_2) = \frac{24,000}{1+\frac{200}{e^{0.35 \cdot CO_2}}} - 100
\label{eq:phimaxco2}
\end{equation}

\begin{center}
  \begin{tabular}{llll}
    \toprule
    Variable      & Description                & Units                 & Source   \\
    \midrule
    $\phi_{max}(CO_2)$ & max $\phi_{N,den}$ for a given respiration rate, assuming high NO$_3$ & g N ha$^{-1}$ d$^{-1}$ & \\
    $CO_2$ & soil heterotrophic respiration rate (excluding root respiration) & kg C ha$^{-1}$ d$^{-1}$ & Equation~\ref{eq:co2} \\
    \bottomrule
  \end{tabular}
\end{center}

Note that in Equation~\ref{eq:co2}, we follow \citet{Parton1996} in assuming that $F_{CO_2}(WFPS)=F_n(WFPS)$ and that $F_{CO_2}(T_s)=F_n(T_s)$. This follows from \citet{Parton1996}'s statement in the last paragraph of page 407: "$S_w$ and $S_t$ are the same functions shown in Figures 2a and 2b, respectively", where Figures 2a and 2b represent $F_n(WFPS)$ and $F_n(T_s)$, respectively.

\begin{equation}
CO_2 = CO_{2,max} \cdot F_{CO_2}(WFPS) \cdot F_{CO_2}(T_s)
\label{eq:co2}
\end{equation}

\begin{center}
  \begin{tabular}{llll}
    \toprule
    Variable      & Description                & Units                 & Source   \\
    \midrule
    $CO_2$ & soil heterotrophic respiration rate (excluding root respiration) & kg C ha$^{-1}$ d$^{-1}$ & \\
    $CO_{2,max}$ & maximum soil respiration rate & kg C ha$^{-1}$ d$^{-1}$ & constant at 80 \\
    $F_{CO_2}(WFPS)$ & effect of water-filled pore space on soil respiration rate & unitless & Equation~\ref{eq:fnwfps} \\
    $F_{CO_2}(T_s)$ & effect of soil temperature on soil respiration rate & unitless & Equation~\ref{eq:fnts} \\
    \bottomrule
  \end{tabular}
\end{center}

Continuing from \citet{Parton1996}:

\begin{equation}
F_d(WFPS) = \frac{a}{b^{\left(\dfrac{c}{b^{(d \cdot WFPS)}}\right)}}
\label{eq:fdwfps}
\end{equation}

\begin{center}
  \begin{tabular}{llll}
    \toprule
    Variable      & Description                & Units                 & Source   \\
    \midrule
    $F_d(WFPS)$ & effect of WFPS on denitrification & unitless & \\
    $a$ & amplitude parameter & unitless & Table~\ref{table:fdwfps-parameters} \\
    $b$ & threshold parameter & unitless & Table~\ref{table:fdwfps-parameters} \\
    $c$ & steepness parameter & unitless & Table~\ref{table:fdwfps-parameters} \\
    $d$ & curvature parameter & unitless & Table~\ref{table:fdwfps-parameters} \\
    \bottomrule
  \end{tabular}
\end{center}

The following table comes from \citet{Parton1996}:

\begin{table}[H]
  \caption{Parameters for $F_d(WFPS)$ in Equation~\ref{eq:fdwfps} based on soil texture}
  \label{table:fdwfps-parameters}
  \centering
  \begin{tabular}{lllll}
    \toprule
    & \multicolumn{4}{c}{Parameter} \\
    \cmidrule(r){2-5}
    Soil Texture & $a$ & $b$ & $c$ & $d$ \\
    \midrule
    sandy & 1.56 & 12.0 & 16.0 & 2.01 \\
    medium & 4.82 & 14.0 & 16.0 & 1.39 \\
    fine & 60.0 & 18.0 & 22.0 & 1.06 \\
    \bottomrule
  \end{tabular}
\end{table}

Unfortunately, \citet{Parton1996} does not explicitly provide a soil texture classification method, i.e., how soil and clay percentages translate into "sandy", "medium", or "fine". As such, we classify soil texture following the USDA broad texture family groupings (coarse/medium/fine) based on sand and clay fractions from the Cycles model configuration, where "coarse" is equivalent to "sandy", "loamy" is equivalent to "medium", and "clayey" is equivalent to "fine". 

The USDA texture classification system uses 12 different categories, so we follow \citet{Ritchey2015}'s convention, in which:

\begin{enumerate}
    \item Sandy soil is soil with 70\% to 100\% sand (this category consists of USDA categories "sand" and "loamy sand")
    \item Fine soil is soil with 35\% to 40\% clay (this category consists of USDA categories "sandy clay", "clay", and "silty clay") 
    \item Medium soil is all other soil that does not fall into the "sandy" or "fine" category (this category consists of USDA categories "sandy clay loam", "sandy loam", "clay loam", "loam", "silty clay loam", "silt loam", and "silt")
\end{enumerate}

Following \citet{DelGrosso2000}:

\begin{equation}
F_r(NO_3/CO_2) = \max(0.16 \cdot k_1,\ k_1 \cdot e^{-0.8 \left( \frac{NO_3}{CO_{2,\mu}} \right)})
\label{eq:frno3co2}
\end{equation}

\begin{center}
  \begin{tabular}{llll}
    \toprule
    Variable      & Description                & Units                 & Source   \\
    \midrule
    $F_r(NO_3/CO_2)$ & NO$_3$-to-respiration ratio factor & unitless & \\
    $NO_3$ & soil NO$_3$ concentration & $\mu$g N gsoil$^{-1}$ & Cycles \\
    $CO_{2,\mu}$ & heterotrophic CO$_2$ respiration & $\mu$g C gsoil$^{-1}$ d$^{-1}$ & Equation~\ref{eq:co2mu} \\
    $k_1$ & maximum N$_2$/N$_2$O ratio (asymptotic ceiling) & unitless & Equation~\ref{eq:k1} \\
    \bottomrule
  \end{tabular}
\end{center}

Note that $NO_3$'s units of "mg N kg$^{-1}$" in Equation~\ref{eq:phimaxno3} and "$\mu$g N gsoil$^{-1}$" in Equation~\ref{eq:frno3co2} are equivalent. However, $CO_{2,\mu}$ from \citet{DelGrosso2000} is not the same as $CO_2$ from \citet{Parton1996}; $CO_{2,\mu}$ (which is measured in $\mu$g C gsoil$^{-1}$ d$^{-1}$) must be derived from $CO_2$ (which is measured in kg C ha$^{-1}$ d$^{-1}$) via the following conversion formula, which relies on bulk density. We follow \citet{DelGrosso2000} in assuming an active soil depth $z$ of 20 cm, due to \citet{Schimel1986}'s observation that most N cycling occurs in the top 5 cm of mineral soil. As \citet{DelGrosso2000} states verbatim on page 1056: "$\cdots$ an active soil depth of 20 cm is assumed because the majority of N cycling occurs in the top 5 cm of mineral soil and N cycling decreases dramatically with depth (\citet{Schimel1986})."

\begin{equation}
CO_{2,\mu} = \frac{CO_2}{10 \cdot BD \cdot z}
\label{eq:co2mu}
\end{equation}

\begin{center}
  \begin{tabular}{llll}
    \toprule
    Variable      & Description                & Units                 & Source   \\
    \midrule
    $CO_{2,\mu}$ & heterotrophic CO$_2$ respiration & $\mu$g C gsoil$^{-1}$ d$^{-1}$ & \\
    $CO_2$ & soil heterotrophic respiration rate (excluding root respiration) & kg C ha$^{-1}$ d$^{-1}$ & Equation~\ref{eq:co2} \\
    $BD$ & bulk density & g cm$^{-3}$ & Cycles \\
    $z$ & soil layer depth & m & constant at 0.20 \\
    \bottomrule
  \end{tabular}
\end{center}

Continuing from \citet{DelGrosso2000}:

\begin{equation}
k_1 = \max(1.7,\ 38.4 - 350 \cdot D_{FC})
\label{eq:k1}
\end{equation}

\begin{center}
  \begin{tabular}{llll}
    \toprule
    Variable      & Description                & Units                 & Source   \\
    \midrule
    $k_1$ & maximum N$_2$/N$_2$O ratio (asymptotic ceiling) & unitless & \\
    $D_{FC}$ & gas diffusivity at field capacity & unitless & Equation~\ref{eq:dfc} \\
    \bottomrule
  \end{tabular}
\end{center}

In order to obtain $D_{FC}$, we must begin by finding $\frac{D_s}{D_0}$.  Following \citet{Potter1996} as in \citet{DelGrosso2000}, we use the formulation of \citet{Millington1971} with polynomial approximations of the exponent terms:

\begin{equation}
\frac{D_s}{D_0} = \frac {(1-S_{WA})^2 \cdot [A- \frac {\Theta_A} {A} + S]^{2z} \cdot [1-P^{2x}] \cdot [(P- \Theta_P) - (P - \Theta_P)^{2y}]} {(1-S_{WA})^2 \cdot [A- \frac {\Theta_A} {A} + S]^{2} \cdot [1-P^{2x}] + (P- \Theta_P) - (P - \Theta_P)^{2y}} + (1-S_{WP})^2 (P-\Theta_P)^{2y}
\label{eq:ds-d0}
\end{equation}

\begin{center}
  \begin{tabular}{llll}
    \toprule
    Variable      & Description                & Units                 & Source   \\
    \midrule
    $\frac{D_s}{D_0}$ & normalized soil gas diffusivity at field capacity & unitless & \\
    $S$ & solid phase volume fraction & m$^3$ m$^{-3}$ & Equation~\ref{eq:s} \\
    $A$ & intra-aggregate pore space & m$^3$ m$^{-3}$ & Equation~\ref{eq:a} \\
    $P$ & inter-aggregate pore space & m$^3$ m$^{-3}$ & Equation~\ref{eq:p} \\
    $\Theta_A$ & volumetric water content in intra-aggregate pores at FC & m$^3$ m$^{-3}$ & Equation~\ref{eq:theta_a} \\
    $\Theta_P$ & volumetric water content in inter-aggregate pores at FC & m$^3$ m$^{-3}$ & Equation~\ref{eq:theta_p} \\
    $S_{WA}$ & fractional liquid saturation of intra-aggregate pore space & unitless & Equation~\ref{eq:s_wa_1},~\ref{eq:s_wa} \\
    $S_{WP}$ & fractional liquid saturation of inter-aggregate pore space & unitless & Equation~\ref{eq:s_wp},~\ref{eq:s_wp_0} \\
    $x$ & tortuosity exponent for inter-aggregate pore space & unitless & Equation~\ref{eq:x} \\
    $y$ & tortuosity exponent for inter-aggregate water content & unitless & Equation~\ref{eq:y} \\
    $z$ & tortuosity exponent for intra-aggregate pore space & unitless & Equation~\ref{eq:z} \\
    \bottomrule
  \end{tabular}
\end{center}

Note that in Equation~\ref{eq:ds-d0}, we follow the literal typesetting of \citet{Potter1996}'s equation, treating $\frac {\Theta_A} {A}$ as a standalone fraction, rather than treating $\frac{A- \Theta_A} {A+S}$ as a fraction. That said, the ambiguity doesn't affect the results of our paper, since the term $(1-S_{WA})^2$ where $S_{WA}=1$ simplifies the entire multiplicative expression to 0 anyway.

Following \citet{Potter1996}'s estimate:

\begin{equation}
S=1-PC
\label{eq:s}
\end{equation}

\begin{center}
  \begin{tabular}{llll}
    \toprule
    Variable      & Description                & Units                 & Source   \\
    \midrule
    $S$ & solid phase volume fraction & m$^3$ m$^{-3}$ & \\
    $PC$ & total pore space capacity (porosity) & m$^3$ m$^{-3}$ & Equation~\ref{eq:pc} \\
    \bottomrule
  \end{tabular}
\end{center}

To find $PC$, we use \citet{Davidson1995}'s assumed particle density value:

\begin{equation}
PC = 1 - \frac{BD}{PD}
\label{eq:pc}
\end{equation}

\begin{center}
  \begin{tabular}{llll}
    \toprule
    Variable      & Description                & Units                 & Source   \\
    \midrule
    $PC$ & total pore space capacity (porosity) & m$^3$ m$^{-3}$ & \\
    $BD$ & bulk density & g cm$^{-3}$ & Cycles \\
    $PD$ & particle density & g cm$^{-3}$ & constant at 2.65 \\ 
    \bottomrule
  \end{tabular}
\end{center}

To find $A$, we follow \citet{Davidson1995}'s lead in equating $A$ with $\theta_{fc}$:

\begin{equation}
A = \theta_{fc}
\label{eq:a}
\end{equation}

\begin{center}
  \begin{tabular}{llll}
    \toprule
    Variable      & Description                & Units                 & Source   \\
    \midrule
    $A$ & intra-aggregate pore space & m$^3$ m$^{-3}$ & \\
    $\theta_{fc}$ & volumetric water content at field capacity & m$^3$ m$^{-3}$ & Equation~\ref{eq:theta_fc} \\
    \bottomrule
  \end{tabular}
\end{center}

We similarly follow \citet{Potter1996} in estimating $P$:

\begin{equation}
P = |PC - A|
\label{eq:p}
\end{equation}

\begin{center}
  \begin{tabular}{llll}
    \toprule
    Variable      & Description                & Units                 & Source   \\
    \midrule
    $P$ & inter-aggregate pore space & m$^3$ m$^{-3}$ & \\
    $PC$ & total pore space capacity (porosity) & m$^3$ m$^{-3}$ & Equation~\ref{eq:pc} \\
    $A$ & intra-aggregate pore space & m$^3$ m$^{-3}$ & Equation~\ref{eq:a} \\
    \bottomrule
  \end{tabular}
\end{center}

To find $\Theta_A$ and $\Theta_P$, we begin with \citet{Davidson1995}'s assumption: "Intra-aggregate porosity was estimated from volumetric soil water content at field capacity \ldots water has drained from the inter-aggregate pore spaces, leaving the intra-aggregate pore spaces 100\% water-filled \ldots it is assumed that the intra-aggregate spaces store water first and lose water last", which forms the basis of the following two equations:

\begin{equation}
S_{WA} = 1
\label{eq:s_wa_1}
\end{equation}

\begin{equation}
\Theta_P = 0
\label{eq:theta_p}
\end{equation}

We also know from \citet{Potter1996} that "$S_{WA}$ and $S_{WP}$ denote the fractional liquid saturation of their respective $A$ and $P$ components of the total void volume", which forms the basis for the following two equations:

\begin{equation}
S_{WA} = \frac{\Theta_A}{A}
\label{eq:s_wa}
\end{equation}

\begin{center}
  \begin{tabular}{llll}
    \toprule
    Variable      & Description                & Units                 & Source   \\
    \midrule
    $S_{WA}$ & fractional liquid saturation of intra-aggregate pore space & unitless & \\
    $\Theta_A$ & volumetric water content in intra-aggregate pores at field capacity & m$^3$ m$^{-3}$ & Equation~\ref{eq:theta_a} \\
    $A$ & intra-aggregate pore space & m$^3$ m$^{-3}$ & Equation~\ref{eq:a} \\
    \bottomrule
  \end{tabular}
\end{center}

\begin{equation}
S_{WP} = \frac{\Theta_P}{P}
\label{eq:s_wp}
\end{equation}

\begin{center}
  \begin{tabular}{llll}
    \toprule
    Variable      & Description                & Units                 & Source   \\
    \midrule
    $S_{WP}$ & fractional liquid saturation of inter-aggregate pore space & unitless & \\
    $\Theta_P$ & volumetric water content in inter-aggregate pores at FC & m$^3$ m$^{-3}$ & Equation~\ref{eq:theta_p} \\
    $P$ & inter-aggregate pore space & m$^3$ m$^{-3}$ & Equation~\ref{eq:p} \\
    \bottomrule
  \end{tabular}
\end{center}

By combining Equation~\ref{eq:s_wa_1} and Equation~\ref{eq:s_wa}, we find that $S_{WA} = \frac{\Theta_A}{A} = 1$; or equivalently, when combined with Equation~\ref{eq:a}:

\begin{equation}
\Theta_A = A = \theta_{fc}
\label{eq:theta_a}
\end{equation}

\begin{center}
  \begin{tabular}{llll}
    \toprule
    Variable      & Description                & Units                 & Source   \\
    \midrule
    $\Theta_A$ & volumetric water content in intra-aggregate pores at field capacity & m$^3$ m$^{-3}$ & \\
    $A$ & intra-aggregate pore space & m$^3$ m$^{-3}$ & Equation~\ref{eq:a} \\
    \bottomrule
  \end{tabular}
\end{center}

By combining Equation~\ref{eq:s_wp} with Equation~\ref{eq:theta_p}, we furthermore find:

\begin{equation}
S_{WP} = 0
\label{eq:s_wp_0}
\end{equation}

\begin{center}
  \begin{tabular}{llll}
    \toprule
    Variable      & Description                & Units                 & Source   \\
    \midrule
    $S_{WP}$ & fractional liquid saturation of inter-aggregate pore space & unitless & \\
    \bottomrule
  \end{tabular}
\end{center}

Finally, rather than solving numerically for $x$, $y$, and $z$ as described by \citet{Millington1971}, we will use \citet{Potter1996}'s polynomial approximations:

\begin{subequations}
\begin{align}
x &= 0.477 P^3 - 0.596 P^2 + 0.437 P + 0.564 \label{eq:x} \\
y &= 0.477 (P - \Theta_P)^3 - 0.596 (P - \Theta_P)^2 + 0.437 (P - \Theta_P) + 0.564 \label{eq:y} \\
z &= 0.477 \left[\frac{A - \Theta_A}{A + S}\right]^3 - 0.596 \left[\frac{A - \Theta_A}{A + S}\right]^2 + 0.437 \left[\frac{A - \Theta_A}{A + S}\right] + 0.564 \label{eq:z}
\end{align}
\end{subequations}

\begin{center}
  \begin{tabular}{llll}
    \toprule
    Variable      & Description                & Units                 & Source   \\
    \midrule
    $x$ & tortuosity exponent for inter-aggregate pore space & unitless & \\
    $y$ & tortuosity exponent for inter-aggregate water content & unitless & \\
    $z$ & tortuosity exponent for intra-aggregate pore space & unitless & \\
    $\Theta_A$ & volumetric water content in intra-aggregate pores at FC & m$^3$ m$^{-3}$ & Equation~\ref{eq:theta_a} \\
    $\Theta_P$ & volumetric water content in inter-aggregate pores at FC & m$^3$ m$^{-3}$ & Equation~\ref{eq:theta_p} \\
    $S$ & solid phase volume fraction & m$^3$ m$^{-3}$ & Equation~\ref{eq:s} \\
    $A$ & intra-aggregate pore space & m$^3$ m$^{-3}$ & Equation~\ref{eq:a} \\
    $P$ & inter-aggregate pore space & m$^3$ m$^{-3}$ & Equation~\ref{eq:p} \\
    \bottomrule
  \end{tabular}
\end{center}

Following \citet{Potter1996}, we use Equation 2 from \citet{Saxton1986} and assume the water potential (i.e., tension) to be 33 kPa for medium to fine textures, and 10 kPa for coarse textures, as in \citet{Papendick1981}. (Note that Table 2 in \citet{Saxton1986} has a known error; as noted in the published erratum \citet{Saxton1986a}, Equation~\ref{eq:B_sx} should not have a standalone term of $g_{sx}(\%S)^2$.)

This results in the equation $\Psi = A_{sx} (\theta_{fc}) ^{B_{sx}}$, which will then rearrange into the following form, which nicely parallels Equation~\ref{eq:swclimit}. Note that soil texture inputs (\%S, \%C) used throughout this section were derived from gSSURGO, SoilGrids, or original site-level publications depending on data availability at each site, as provided by the Cycles model developers.

\begin{subequations}
\begin{align}
\theta_{fc} &= \left(\dfrac{\Psi}{A_{sx}}\right)^{\frac{1}{B_{sx}}} \label{eq:theta_fc} \\
A_{sx} &= 100 \cdot \exp\left(a_{sx} + b_{sx}(\%C) + c_{sx}(\%S)^2 + d_{sx}(\%S)^2(\%C)\right) \label{eq:A_sx} \\
B_{sx} &= e_{sx} + f_{sx}(\%C)^2 + g_{sx}(\%S)^2(\%C) \label{eq:B_sx} 
\end{align}
\end{subequations}

\begin{center}
  \begin{tabular}{llll}
    \toprule
    Variable     & Description                 & Units                 & Source   \\
    \midrule
    $\theta_{fc}$ & volumetric water content at field capacity & m$^3$ m$^{-3}$ & \\
    $A_{sx}$ & air entry potential scaling parameter & kPa & \\
    $B_{sx}$ & pore size distribution shape parameter & unitless & \\
    $\Psi$ & water potential (tension) & kPa & Table~\ref{table:theta_fc-parameters} \\
    $\%S$ & percentage of soil that is sand & unitless & site survey \\
    $\%C$ & percentage of soil that is clay & unitless & site survey \\
    $a_{sx}$ & regression coefficient & unitless & constant at -4.396 \\
    $b_{sx}$ & regression coefficient & unitless & constant at -0.0715 \\
    $c_{sx}$ & regression coefficient & unitless & constant at -4.880 $\times$ 10$^{-4}$ \\
    $d_{sx}$ & regression coefficient & unitless & constant at -4.285 $\times$ 10$^{-5}$ \\
    $e_{sx}$ & regression coefficient & unitless & constant at -3.140 \\
    $f_{sx}$ & regression coefficient & unitless & constant at -2.22 $\times$ 10$^{-3}$ \\
    $g_{sx}$ & regression coefficient & unitless & constant at -3.484 $\times$ 10$^{-5}$ \\
    \bottomrule
  \end{tabular}
\end{center}

The following table comes from \citet{Potter1996}'s assumptions based on \citet{Papendick1981}:

\begin{table}[H]
  \caption{Parameters for $\Psi$ in Equation~\ref{eq:theta_fc} based on soil texture}
  \label{table:theta_fc-parameters}
  \centering
  \begin{tabular}{ll}
    \toprule
    Soil Texture & Water Potential $\Psi$ (kPa) \\
    \midrule
    sandy & 10.0 \\
    medium & 33.0 \\
    fine & 33.0 \\
    \bottomrule
  \end{tabular}
\end{table}

Finally, $D_{FC}$ is obtained by evaluating Equation~\ref{eq:ds-d0} at $\theta = \theta_{fc}$, following the assumption of \citet{Davidson1995} that intra-aggregate pore spaces are fully water-filled at field capacity ($\Theta_A = \theta_{fc}$, $S_{WA} = 1$) and inter-aggregate pore spaces are air-filled ($\Theta_P = 0$, $S_{WP} = 0$):

\begin{equation}
D_{FC} = \left.\frac{D_s}{D_0}\right|_{\Theta_A = \theta_{fc},S_{WA}=1,\Theta_P=0,S_{WP}=0}
\label{eq:dfc}
\end{equation}

Then, continued from \citet{DelGrosso2000}, we use \citet{Hartman2019}'s convention of treating $WFPS$ as the average across the 2nd and 3rd soil layers specifically. Note that we multiply $WFPS$ by 100; this doesn't appear in the original equation from \citet{DelGrosso2000}, but is necessary for us since we define $WFPS$ as a fraction, whereas $WFPS$ is treated as a percent in the original equation.

\begin{equation}
F_r(WFPS) = 
\begin{cases}
\max(0.1,0.015(100 \cdot WFPS) - 0.32) & \text{if soil is intact} \\[10pt]
\max(0.1,0.05(100 \cdot WFPS) - 3.1) & \text{if soil is repacked}
\end{cases}
\label{eq:frwfps} 
\end{equation}

\begin{center}
  \begin{tabular}{llll}
    \toprule
    Variable      & Description                & Units                 & Source   \\
    \midrule
    $F_r(WFPS)$ & effect of water-filled pore space on $R_{N_2/N_2O}$ & unitless & \\
    $WFPS$ & weighted average water-filled pore space in 2nd and 3rd soil layers & unitless & Equation~\ref{eq:wfps} \\
    \bottomrule
  \end{tabular}
\end{center}

As explained in Section~\ref{subsec:dataset}, we determined site-specific soil intactness designations (intact vs. repacked) from the historical tillage records available for each of our sites.

\subsection{Additional Figures}
\label{subsec:additional-figures}

\begin{figure}[h]
    \centering
    \includegraphics[width=0.9\linewidth]{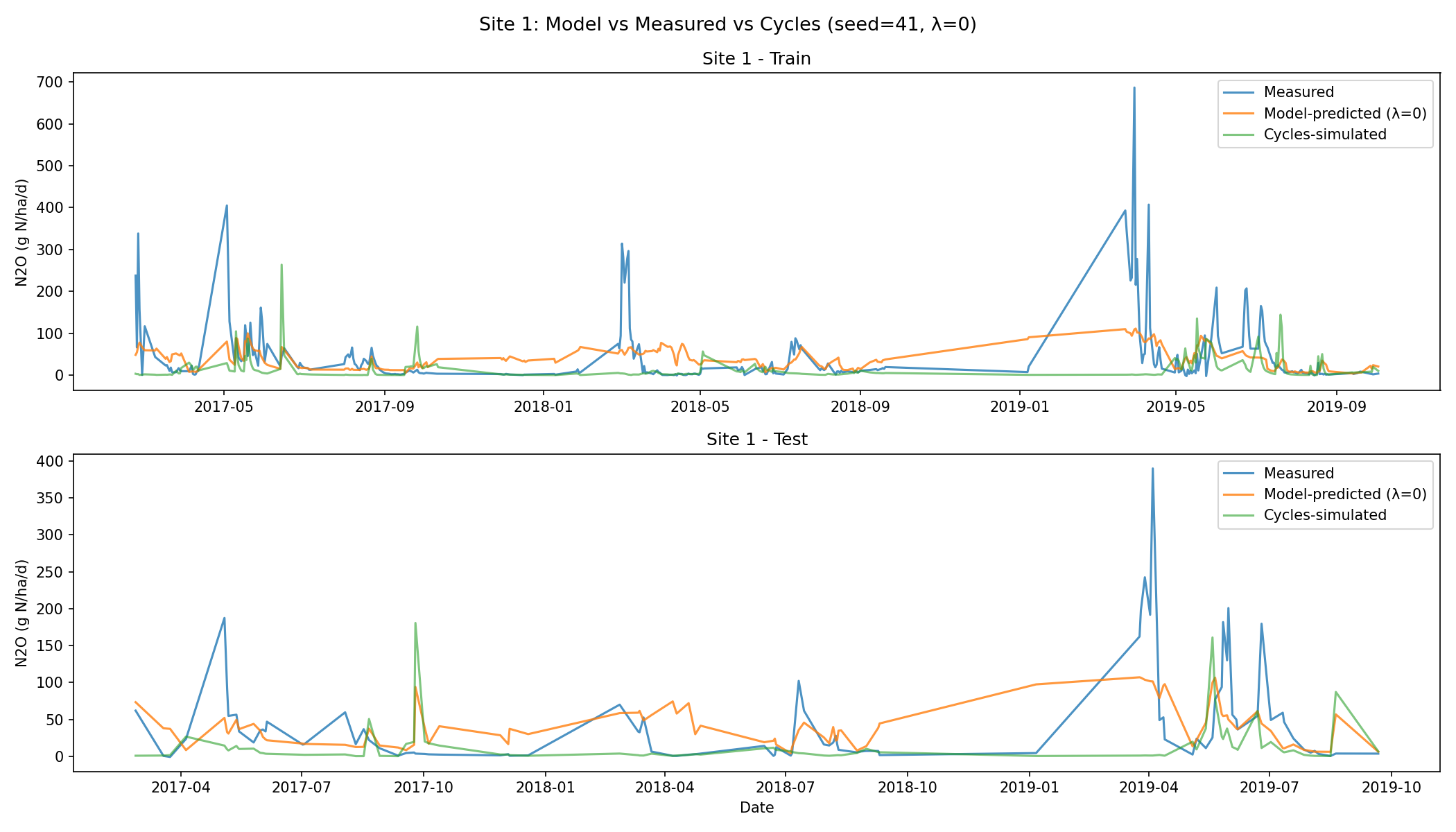}
    \caption{Site 1 complete time series}
    \label{fig:site-1}
\end{figure}

\begin{figure}[h]
    \centering
    \includegraphics[width=0.9\linewidth]{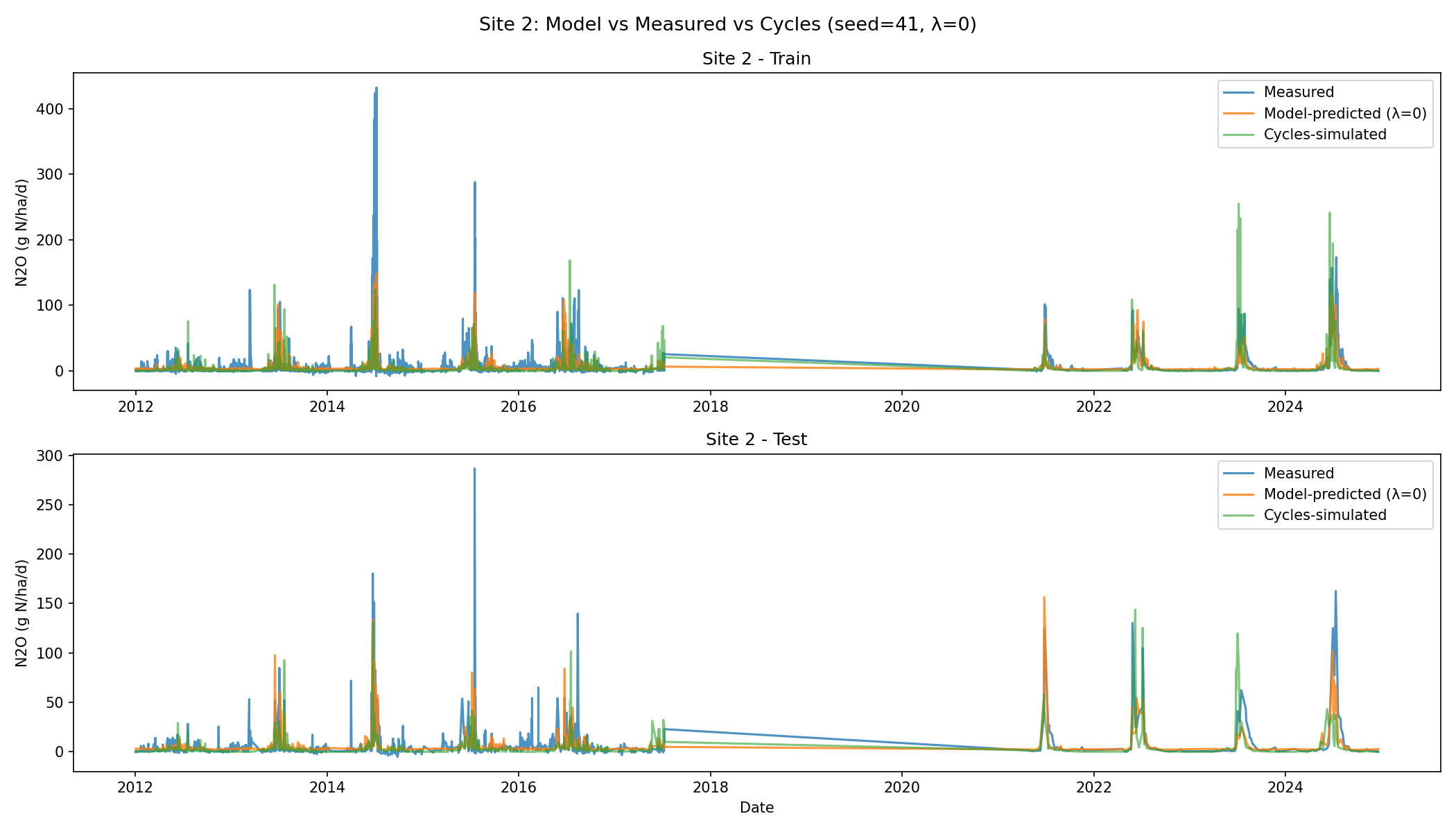}
    \caption{Site 2 complete time series}
    \label{fig:site-2}
\end{figure}

\begin{figure}[h]
    \centering
    \includegraphics[width=0.9\linewidth]{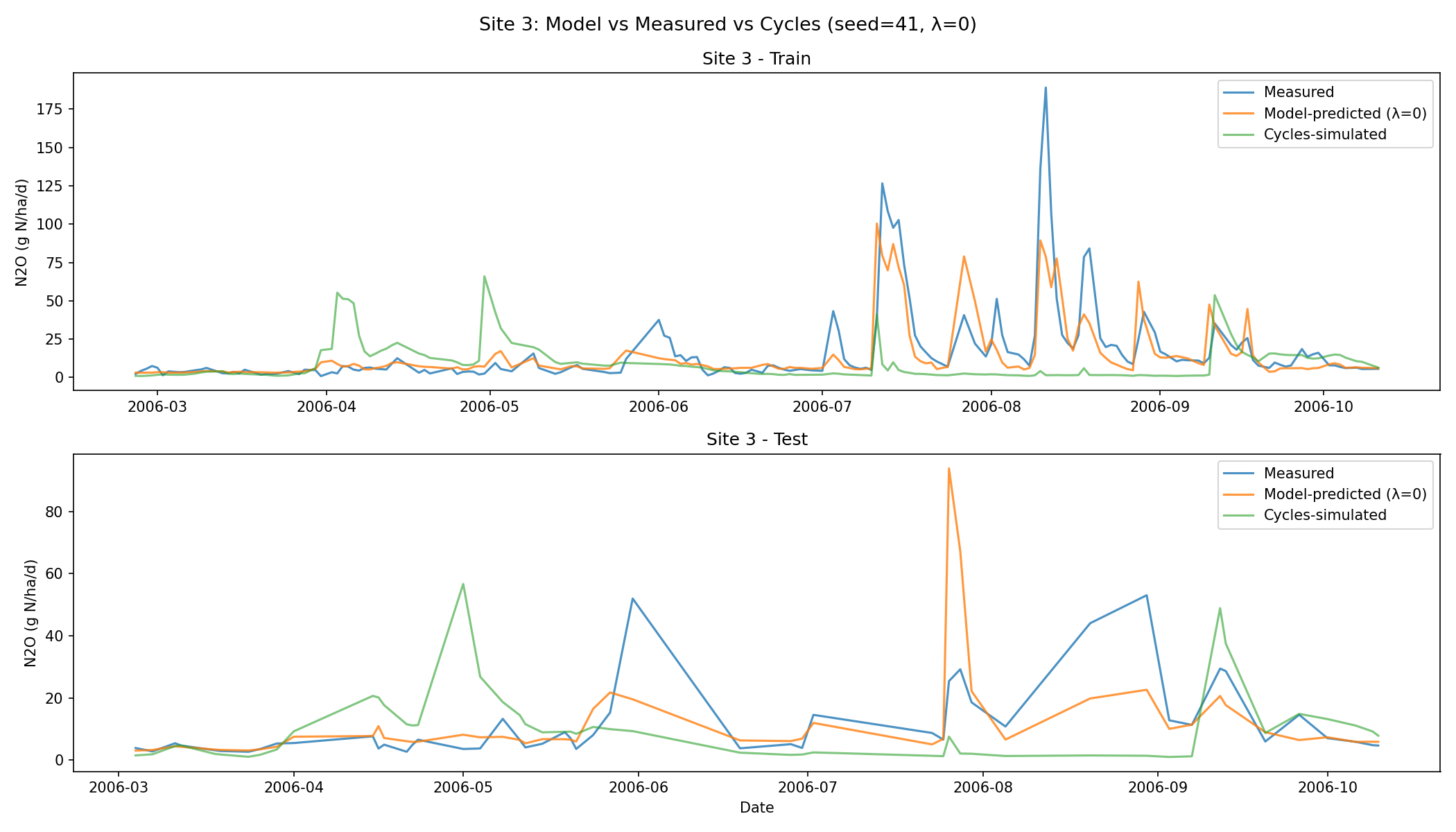}
    \caption{Site 3 complete time series}
    \label{fig:site-3}
\end{figure}

\begin{figure}[h]
    \centering
    \includegraphics[width=0.9\linewidth]{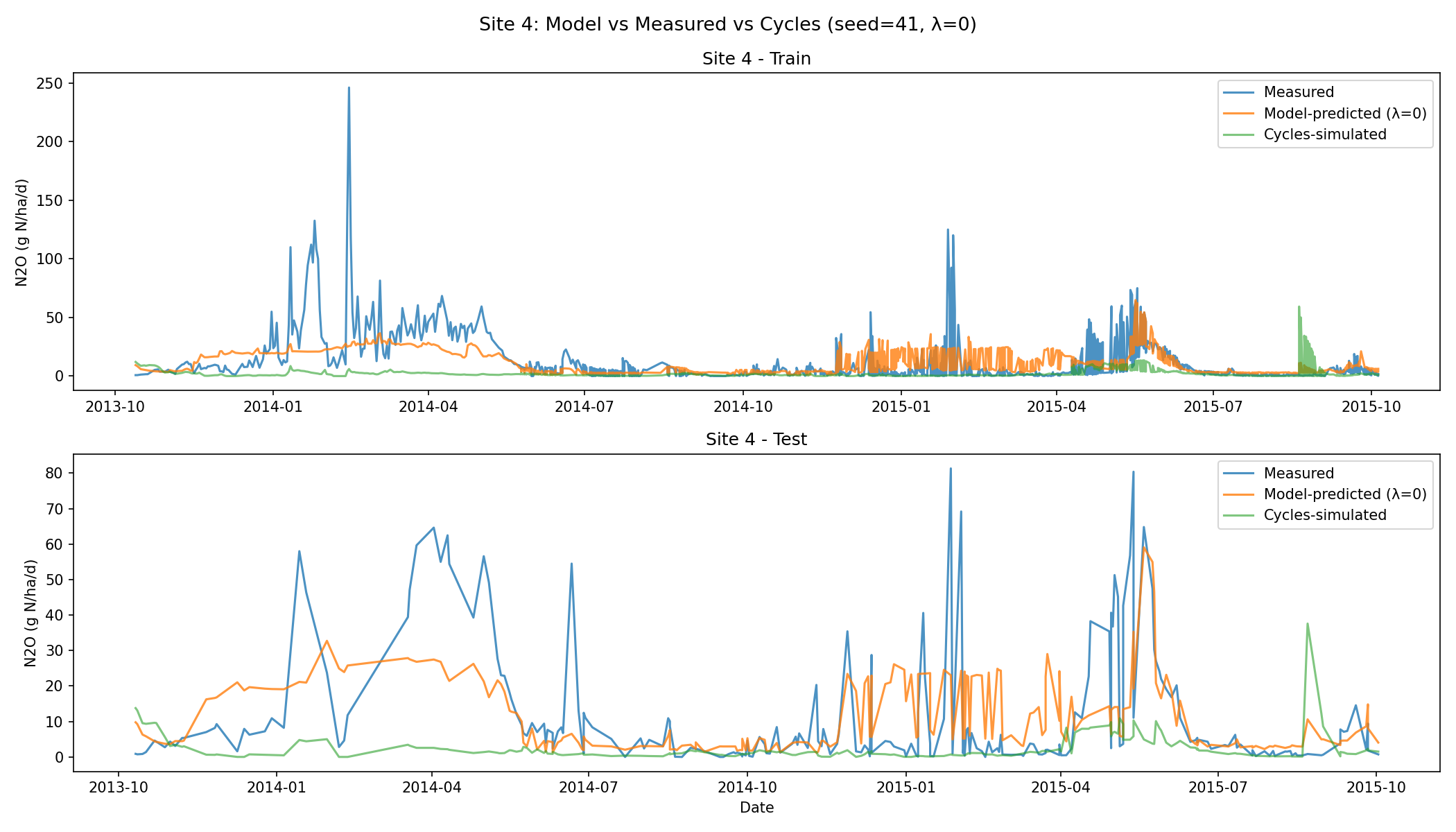}
    \caption{Site 4 complete time series}
    \label{fig:site-4}
\end{figure}

\subsection{Additional Tables}
\label{subsec:additional-tables}

\begin{sidewaystable}
  \caption{R$^2$ across $\lambda$ sweep (holdout validation)}
  \label{table:lambda-sweep-holdout}
  \centering
  \begin{tabular}{llllllllllllll}
    \toprule
    lambda & 0 & 0.01 & 0.05 & 0.1 & 0.2 & 0.3 & 0.4 & 0.5 & 0.6 & 0.7 & 0.8 & 0.9 & 1 \\
    \midrule
    seed & ~ & ~ & ~ & ~ & ~ & ~ & ~ & ~ & ~ & ~ & ~ & ~ & ~ \\
    \midrule
    41 & 0.442 & 0.479 & 0.439 & 0.449 & 0.358 & 0.322 & 0.274 & 0.227 & 0.166 & 0.16 & 0.19 & 0.173 & 0.135 \\
    42 & 0.44 & 0.438 & 0.435 & 0.432 & 0.402 & 0.376 & 0.358 & 0.306 & 0.3 & 0.275 & 0.242 & 0.235 & 0.188 \\
    43 & 0.379 & 0.338 & 0.391 & 0.346 & 0.28 & 0.233 & 0.168 & 0.133 & 0.019 & -0.043 & -0.06 & -0.182 & -0.224 \\
    44 & 0.412 & 0.421 & 0.386 & 0.369 & 0.335 & 0.291 & 0.187 & 0.159 & 0.14 & 0.106 & 0.112 & 0.091 & 0.113 \\
    45 & 0.45 & 0.457 & 0.434 & 0.459 & 0.419 & 0.402 & 0.375 & 0.364 & 0.299 & 0.244 & 0.234 & 0.25 & 0.189 \\
    46 & 0.395 & 0.399 & 0.399 & 0.402 & 0.382 & 0.368 & 0.343 & 0.312 & 0.263 & 0.254 & 0.161 & 0.191 & 0.168 \\
    47 & 0.386 & 0.354 & 0.318 & 0.344 & 0.284 & 0.24 & 0.191 & 0.161 & 0.115 & 0.115 & 0.081 & 0.052 & 0.055 \\
    48 & 0.399 & 0.396 & 0.387 & 0.374 & 0.359 & 0.312 & 0.267 & 0.22 & 0.169 & 0.11 & 0.093 & 0.074 & 0.037 \\
    49 & 0.43 & 0.439 & 0.434 & 0.431 & 0.409 & 0.369 & 0.343 & 0.288 & 0.258 & 0.219 & 0.192 & 0.153 & 0.099 \\
    50 & 0.374 & 0.343 & 0.338 & 0.351 & 0.299 & 0.25 & 0.23 & 0.156 & 0.155 & 0.148 & 0.144 & 0.139 & 0.145 \\
    \midrule
    mean & 0.411 & 0.406 & 0.396 & 0.396 & 0.353 & 0.316 & 0.274 & 0.233 & 0.188 & 0.159 & 0.139 & 0.118 & 0.09 \\
    std & 0.027 & 0.047 & 0.04 & 0.042 & 0.049 & 0.058 & 0.074 & 0.077 & 0.086 & 0.09 & 0.084 & 0.118 & 0.116 \\
    \bottomrule
  \end{tabular}
\end{sidewaystable}

\begin{sidewaystable}
  \caption{R$^2$ across $\lambda$ sweep (leave-one-site-out validation)}
  \label{table:lambda-sweep-loso}
  \centering
  \begin{tabular}{llllllllllllll}
    \toprule
    lambda & 0 & 0.01 & 0.05 & 0.1 & 0.2 & 0.3 & 0.4 & 0.5 & 0.6 & 0.7 & 0.8 & 0.9 & 1 \\
    \midrule
    seed & ~ & ~ & ~ & ~ & ~ & ~ & ~ & ~ & ~ & ~ & ~ & ~ & ~ \\
    \midrule
    41 & -12.466 & -7.126 & -60.352 & -16.591 & -19.251 & -11.576 & -16.524 & -3.431 & -6.179 & -4 & -4.233 & -3.729 & -5.936 \\
    42 & -18.642 & -25.336 & -30.724 & -7.637 & -23.205 & -32.59 & -7.347 & -9.48 & -2.452 & -3.247 & -4.539 & -3.769 & -6.531 \\
    43 & -180.562 & -78.535 & -29.245 & -1.539 & -27.242 & -203.175 & -129.547 & -80.989 & -101.11 & -21.849 & -56.527 & -37.617 & -47.983 \\
    44 & -11.944 & -39.53 & -17.288 & -28.272 & -45.996 & -11.974 & -3.589 & -3.814 & -3.265 & -16.556 & -4.855 & -4.87 & -5.273 \\
    45 & -85.602 & -62.964 & -33.413 & -26.55 & -87.985 & -45.606 & -5.258 & -31.819 & -24.96 & -32.353 & -34.366 & -4.598 & -41.846 \\
    46 & -48.069 & -0.883 & -1.529 & -115.415 & -77.687 & -12.285 & -2.369 & -13.605 & -23.389 & -3.294 & -3.705 & -5.343 & -7.065 \\
    47 & -74.384 & -118.931 & -32.469 & -65.23 & -47.697 & -102.69 & -60.357 & -22.196 & -39.976 & -30.938 & -42.735 & -53.996 & -40.265 \\
    48 & -63.324 & -95.2 & -77.864 & -42.524 & -5.018 & -19.809 & -19.711 & -25.817 & -8.816 & -4.218 & -5.629 & -5.091 & -6.625 \\
    49 & -147.117 & -311.641 & -136.065 & -101.067 & -109.851 & -30.654 & -19.079 & -6.473 & -4.287 & -6.084 & -3.447 & -4.956 & -4.214 \\
    50 & -74.608 & -6.746 & -92.948 & -33.315 & -46.036 & -31.805 & -10.782 & -6.218 & -4.532 & -3.43 & -1.873 & -5.343 & -6.353 \\
    \midrule
    mean & -71.672 & -74.689 & -51.190 & -43.814 & -48.997 & -50.216 & -27.456 & -20.384 & -21.897 & -12.597 & -16.191 & -12.931 & -17.209 \\
    std & 53.170 & 87.736 & 38.633 & 36.540 & 31.660 & 57.028 & 37.588 & 22.266 & 28.928 & 11.281 & 19.246 & 16.849 & 17.236 \\
    \bottomrule
  \end{tabular}
\end{sidewaystable}


\newpage

\end{document}